%% file: hybridpso.tex
\RequirePackage{letltxmacro}
\LetLtxMacro{\LaTeXtextbf}{\textbf}
\documentclass{ieeeaccess}
\LetLtxMacro{\textbf}{\LaTeXtextbf}

\usepackage{cite}
\usepackage{amsmath,amssymb,amsfonts}
\usepackage{algorithmic}
\usepackage{graphicx}
\usepackage{textcomp}

\usepackage[flushleft]{threeparttable}
\usepackage{stfloats}

\usepackage{multirow}
\usepackage{hhline}

\usepackage[linesnumbered,lined,boxed,commentsnumbered,ruled,longend]{algorithm2e}

\usepackage{adjustbox,lipsum}

\def\BibTeX{{\rm B\kern-.05em{\sc i\kern-.025em b}\kern-.08em
    T\kern-.1667em\lower.7ex\hbox{E}\kern-.125emX}}
\begin{document}
\doi{10.1109/ACCESS.2017.DOI}

\title{PSO-Convolutional Neural Networks with Heterogeneous Learning Rate}
\author{
\uppercase{Nguyen Huu Phong}\authorrefmark{1},
\uppercase{Augusto Santos\authorrefmark{2}, and Bernardete~Ribeiro}\authorrefmark{3},\IEEEmembership{Senior Member, IEEE}}
\address[1,2,3]{CISUC, Department of Informatics Engineering, University of Coimbra, Coimbra Portugal}

\tfootnote{``This research is sponsored by FEDER funds through the programs COMPETE -- ``Programa Operacional Factores de Competitividade'' and Centro2020 -- ``Centro Portugal Regional Operational Programme'', and by national funds through FCT -- ``Funda\c{c}\~{a}o para a Ci\^encia e a Tecnologia'', under the project UIDB/00326/2020.''
}


\corresp{Corresponding author: Nguyen Huu Phong (e-mail: phong@dei.uc.pt).}
\begin{abstract}
 Convolutional Neural Networks (ConvNets or CNNs) have been candidly deployed in the scope of computer vision and related fields. Nevertheless, the dynamics of training of these neural networks lie still elusive: it is \emph{hard} and computationally expensive to train them. A myriad of architectures and training strategies have been proposed to overcome this challenge and address several problems in image processing such as speech, image and action recognition as well as object detection. In this article, we propose a novel Particle Swarm Optimization (PSO) based training for ConvNets. In such framework, the vector of weights of each ConvNet is typically cast as the position of a particle in phase space whereby PSO collaborative dynamics intertwines with Stochastic Gradient Descent (SGD) in order to boost training performance and generalization. Our approach goes as follows: $i)$ [\textbf{regular phase}] each ConvNet is trained independently via SGD; $ii)$ [\textbf{collaborative phase}] ConvNets share among themselves their current vector of weights (or particle-position) along with their gradient estimates of the Loss function. Distinct step sizes are coined by distinct ConvNets. By properly blending ConvNets with large (possibly random) step-sizes along with more conservative ones, we propose an algorithm with competitive performance with respect to other PSO-based approaches on Cifar-10 and Cifar-100 (accuracy of $98.31\%$ and $87.48\%$). These accuracy levels are obtained by resorting to only four ConvNets -- such results are expected to scale with the number of collaborative ConvNets accordingly. We make our source codes available for download https://github.com/leonlha/PSO-ConvNet-Dynamics.
\end{abstract}
\titlepgskip=-15pt
\maketitle
\section{Introduction}
 \IEEEPARstart{I}{m}age classification lies at the core of many computer vision related tasks: extracting relevant information from telescope images in astronomy, navigation in robotics, cancer classification in medical image, security, to cite a few examples. It is currently almost an ontological commitment that Convolutional Neural Networks are particularly tailored to Image Processing. However, training these (large-scale) neural networks for generalization is still a big part of the puzzle since the performance is sensitive to the architecture, the training set, the sample size (a.k.a. sample complexity) among other attributes.

More concretely, in the classification problem, ConvNets -- or, more broadly, Deep Neural Networks -- are trained to learn a target function $f^{\star}\,:\,\mathbb{R}^{N}\longrightarrow \left\{0,1,2,\ldots,M\right\}$ traditionally via Stochastic Gradient Descent (SGD)
\begin{equation}\label{eq:SGD}
\mathbf{\theta}(t+1)=\mathbf{\theta}(t)-\frac{\eta}{K}\sum\limits_{i=1}^K\nabla_{\theta}\widehat{L}\left(x_i,f^{\star}(x_i),\theta(t)\right),
\end{equation}
where~$\widehat{L}$ is an estimate of the loss function; $\theta(t)$ is the vector collecting all the weights of the neural network at the iterate~$t$; $\eta$ is the step size; $K<T$ is the batch size and~$\left\{x_i,f^{\star}(x_i)\right\}_{i=1}^T$ is the training set. The landscape being high-dimensional, nonconvex and with a \emph{geometry}\footnote{The term geometry hereby subsumes a collection of attributes of the Loss function, e.g., regularity (how smooth it is), how deep and wide are the minima, etc., that impacts quite critically the SGD dynamics~\eqref{eq:SGD}.} that is sensitive to the architecture render the problem quite unstable to tuning. Generalization and consistency may be technically studied in certain ideal cases, e.g., under certain thermodynamic limit regimes on the number of neurons~\cite{mei2018mean}. But for practical purposes, ascertaining a \emph{proper} structure for the network -- i.e., a parsimonious structure yielding a Loss function that is amenable to generalization -- is still quite challenging.

Recently, distributed approaches have been leveraged to overcome the \emph{geometry} sensitivity of the landscape and yield a more robust approach to generalization. At the limelight lies nature inspired approaches: Particle Swarm Optimization (PSO)~\cite{houssein2021major,kennedy1995particle,shi1998modified}, Ant Colony Optimization (ACO)~\cite{rokbani2021bi,dorigo2006ant}, Artificial Bee Colony Optimization (ABCO)~\cite{jacob2021artificial,karaboga2005idea}, etc. The core idea is that each \emph{particle} treads the landscape exchanging information with neighboring particles about its current estimate of the geometry (e.g., the gradient of the Loss function) and its position. The overall goal in this framework is to devise a distributed collaborative algorithm that boosts the optimization performance by leading (at least some of the) particles up to the \emph{best} minimum.

In this work, we propose a modified PSO-ConvNet training by incorporating some elements of Cucker-Smale dynamics~\cite{cucker2007emergent} into the PSO algorithm. In addition, a novel alternative dynamics is introduced based on our observation of positions of particles. Further, we set one of the particles with a large random step-size. The idea in its core is simple: i) this \emph{wilder} particle can scan the land in a faster time-scale; ii) it can only be trapped by \emph{deeper} minima; iii) by properly tuning the weights, we enable this particle to attract the more conservative ones to the stronger minimum. A more detailed description of the algorithm will be provided in Section~\ref{sec:proposed}.

The main contributions of this paper are:
\begin{enumerate}   
\item It is the first work to integrate a modified Cucker-Smale dynamics into PSO algorithm on image classification to enhance particle's exploration capability.  
\item A novel dynamics is proposed that achieves state-of-the-art results.
\item Special characteristics of hybrid ConvNet-PSO is taken into consideration by introducing wilder particles incorporated by a restricted random learning rate.
\item Extensive experiments are performed on large-scale Cifar-10 and Cifar-100 benchmark datasets to verify the effectiveness of the proposed methods
\end{enumerate}  

The rest of the paper is organized as follows: In Section \ref{sec:background}, we discuss backgrounds of Convolution Neural Networks and Particle Swarm Optimization to clarify our stand. In addition, we summarize recent advances in swarm intelligent algorithms for optimization of ConvNet's hyper-parameters and tuning ConvNet's learning rate. In Section~\ref{sec:proposed}, we describe our approach in detail from its dynamics formulas to proposed methods for improvement. Experiments and result discussions are presented in Section~\ref{sec:experiments} and ~\ref{sec:results}, respectively. In Section~\ref{sec:stateoftheart}, we compare our approach with recent state-of-the-art to claim its superiority. Ultimately, we conclude our work in Section~\ref{sec:conclusion}.
\section{Background and Related Works}
\label{sec:background}
In this section, we discuss backgrounds of convolutional neural networks and particle swarm optimization as well as relevant works of hybrid PSO-ConvNets and learning rate.
\subsection{Background}
\subsubsection{Convolutional Neural Networks}
ConvNets have demonstrated superior performance when compared to other methods in computer vision and related tasks. The technique was originally proposed by LeCun et al.~\cite{lecun1989backpropagation} in 1989, though, only after 2012 when AlexNet~\cite{krizhevsky2012imagenet} outperformed contemporary state-of-the-art, ConvNets became the most representative neural networks in the area. This breakthrough in popularity has been motivated largely by: (i) availability of high-performance computing hardware—particularly, modern graphical processing units (GPUs) and (ii) promotions of large-scale datasets such as the ImageNet Large-Scale Visual Recognition Challenge (ILSVRC). The early designs of ConvNets were shallow and included only a few layers; however, the field has been evolving to cope with the rampant scaling in computer power and image/data resolution. For example, GoogLeNet~\cite{szegedy2015going} -- which was a winner of ILSVRC 2014, introduced Inception model which helps to reduce the computational cost as it reduces significantly the number of parameters involved in a network. VGGNet~\cite{simonyan2014very} in 2014 showed that deeper layers improve the performance of ConvNets. ResNet~\cite{he2016deep} which was the best of ILSVRC in 2015, introduced the idea of residual learning. The later developments involved a delicate trade-off among network depth, width and image resolution as EfficientNet~\cite{tan2019efficientnet} or adaptation for small size devices (MobileNet~\cite{howard2017mobilenets}). In addition, SqueezeNet~\cite{iandola2016squeezenet}, SENet~\cite{hu2018squeeze}, DenseNet~\cite{iandola2014densenet}, ResNeXt~\cite{xie2017aggregated}, Xception~\cite{chollet2017xception} and other families of neural networks thereof, have been proposed and demonstrated to efficiently perform in many applications.

\begin{figure} [hbt!]
\begin{center}
\includegraphics[keepaspectratio,width=0.45\textwidth]{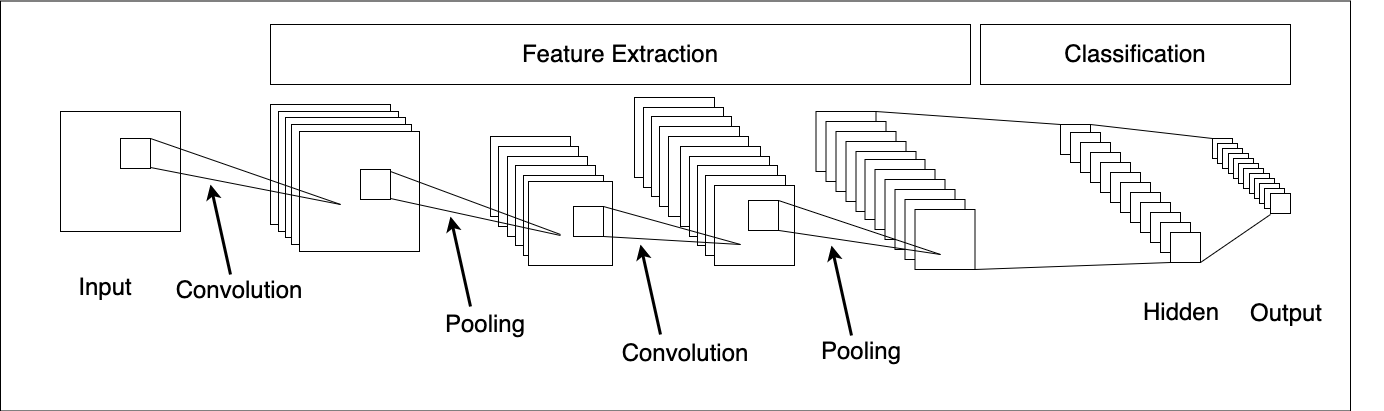}
\caption{Illustration of a traditional ConvNet design~\cite{lecun2010convolutional}.}
\label{fig:convnet}
\end{center}
\end{figure}

As discussed above, ConvNets differ from one to another in their architectures, though, a typical design is illustrated in Figure~\ref{fig:convnet} which stacks multiple trainable layers. The feature extraction module is comprised of several convolution operations for filtering and pooling operations for sub-sampling (reducing the network size). Within this module, the input and output of each layer are given by arrays and referred to as feature maps. The classification module contains fully connected (FC) layers.

For the sake of clarity regarding the operations comprised by a ConvNet, the mathematical formulations can be summarized as follows:
\begin{align}
& \begin{cases}
      O_{i}=X & \text{if $i=1$}\\
      Y_{i}=f_{i}(O_{i-1},W_{i}) & \text{if $i>1$}\\
      O_{i}=g_{i}(Y_{i})
  \end{cases} \\
& \begin{cases}
      Y_{i}=W_{i} \circledast O_{i-1} & \text{$i^{th}$ layer is a convolution}\\
      Y_{i}=\boxplus_{n,m} O_{i-1} & \text{$i^{th}$ layer is a pool}\\
      Y_{i}=W_{i}*O_{i-1} & \text{$i^{th}$ layer is a FC}      
  \end{cases}         
\end{align}
where $X$ represents the input image; $O_{i}$ is the output for layer $i^{th}$; $W_{i}$ indicates the weights of the layer; $f_{i}(\cdot)$ denotes weight operation for convolution, pooling or FC layers; $g_{i}(\cdot)$ is an activation function, for example, sigmoid, tanh and rectified linear (ReLU) or more recently Leaky ReLU~\cite{maas2013rectifier}; The symbol ($\circledast$) acts as a convolution operation which uses \textit{shared} weights to reduce expensive matrix computation~\cite{lecun2010convolutional}; Window ($\boxplus_{n,m}$) shows an average or a max pooling operation which compute average or max values over neighbor region of size $n \times m$ in each feature map. Matrix multiplication of weights between layer $i^{th}$ and the layer $(i-1)^{th}$ in FC is represented as ($*$).
\subsubsection{Particle Swarm Optimization}
Particle Swarm Optimization (PSO) is a population based stochastic optimization algorithm originally introduced by Kennedy and Eberhart in 1995~\cite{kennedy1995particle} and modified in~\cite{shi1998modified}. The attractive feature of PSO is attributed to the ability of particles to learn from others (social behavior) and from their individual experience (cognitive behavior). In PSO, the position of each particle represents a potential solution to a problem and is obtained via random search. At first, each particle (among $N$ independent particles in a D-dimensional phase space) is randomly pre-assigned to a position $x$ in a search space $\Omega^D$, and during the evolution process, continues to update its position according to the following dynamics:
\begin{gather}
\label{eq:PSO_original}
v_{id}(t+1)=wv_{id}(t)+c_1r_1(P_{id}(t)-x_{id}(t))
\notag\\+c_2r_2(P_{gd}(t)-x_{id}(t)),\label{eq:sub1_PSO_original}\nonumber
\\
x_{id}(t+1)=x_{id}(t)+v_{id}(t+1).\label{eq:sub2_PSO_original}
\end{gather}
%
%
%
%
%
%
where, $v_{id}(t)$ and $x_{id}(t)$ represent the $d$ component of the velocity and position of particle $i$ at the iteration (or time) $t$; $r_1$ and $r_2$ are independent random variables uniformly distributed over the interval $\left[0,1\right]$; $P_{id}$ and $P_{gd}$ serve as the particle's own best position and swarm's best position; The $t$ denotes the iteration; The parameter $w$ is called ``inertia weight'' and it was introduced in the modified version besides $c_1$ (social coefficient accelerator) and $c_2$ (cognitive coefficient accelerator) for controlling the behavior of the particles and balancing the interplay between exploration and exploitation. 
PSO is drawn from a simple nature inspired concept with affordable implementation and computational efficiency, even though, the algorithm tends to be trapped into local minima after few iterations. Thus, a great amount of research has been attempting to tackle the problem. For example, authors proposed a PSO algorithm with an adaptive learning strategy (PSO-ALS) that groups a swarm into several sub-swarms to maintain population diversity~\cite{zhang2020particle}. Other authors attempted to tune the inertia-related parameters of the PSO dynamics in order to hinder early convergence to local minima~\cite{han2018APSO}. The inertia related weights were also the primary focus of the works~\cite{LI2019a,bansal2011inertia,NICKABADI2011a,liu2016an}.
\subsection{Related Works}
\subsubsection{Hybrid PSO-ConvNets}
At present, several research studies have been applied to optimizing hyper-parameters of convolutional neural networks. Some works aim at a narrower sense where the hyper-parameters are manually set based on trial-and-error experiments. Others follow a broader sense in which the learning rate and the structure of the layers can be automatically generated. For example, in ~\cite{real2017large}, starting from a simple neural network of one layer, the model evolves into full architectures that are competitive with the state-of-the-art counterparts. Further examples are: i) Evolving Deep CNN (EvoCNN) where the layers are optimized via a genetic algorithm~\cite{sun2019evolving}; ii) genetic programming to optimize the architecture of the CNN for image recognition~\cite{suganuma2017genetic}. Therefore, instead of an arbitrary manual design, evolutionary computation (EC) algorithms have shown their potential in drawing optimal architectures with global optimization search capability.

However, training a ConvNet is computationally expensive, for instance, models in~\cite{real2017large,real2019regularized} critically resort to a cluster of hundreds or even thousands of GPUs which are hardly affordable for most of research centres. In this sense, PSO is a better choice as the algorithm requires much less computation. In contrast to EC methods which evolve via competition, in PSO, particles cooperate to share information, e.g., best position, current location and direction. As demonstrated in~\cite{tu2021modpso}, the fusion of modified particle swarm optimization (ModPSO) together with back-propagation (BP) and convolution neural network was proposed. The method offers strategies to adjust inertia weight, accelerator parameters, update velocity. The involvement of ModPSO and BP improves training ConvNet by avoiding early convergence and local minima. In addition, the dynamic and adaptive parameters strike a trade-off between the global and local search ability, and the diversity of the swarm.

Other works~\cite{wang2019cpso,da2018convolutional} utilize PSOs to search for hyper-parameters of ConvNets. For example, a novel variant of the PSO algorithm (cPSO-CNN) was proposed in ~\cite{wang2019cpso} to optimize the kernel size, filter number, stride and padding. The method improved the PSO’s exploration ability by applying confidence function defined by a normal compound distribution and reformulated PSO’s scalar acceleration coefficients as vectors for better adapting to the variant ranges of the hyper-parameters. In addition, a linear prediction model predicted the ranking of the PSO particles for fast fitness evaluation.

While there have been extensive research on the optimization of ConvNets, the problem is still far from being well understood.
\subsubsection{Learning Rate Tuning}
\label{sec:learningrateworks}
When training ConvNets, learning rate might be the most essential hyperparameter as emphasized by Yoshua Bengio in the practical book ``Neural networks: Tricks of the trade"~\cite{bengio2012practical}. Take into account the kernel size, for example, a subtle image may be convoluted better with a small kernel size, but a larger kernel size would scan just a little worse than its counterpart~\cite{tomen2021spectral}. The larger the number of filters, the better the emergent patterns, though, fewer filters should not cause a significant lost in ConvNet's performance. Setting a learning rate is totally different as a small/large learning rate could compromise convergence in training.

The main objective of learning rate is tuning to find global minima, local minima, or generally an area where loss function obtains adequately low values (ideally the cost reaches zero $L_{(z,\theta)} \rightarrow 0$). Tremendous efforts have been made to reduce execution time to yield better performance.

The goal of learning rate schedules is to regulate the learning rate following a prefixed schedule, e.g., time-based decay, step decay and exponential decay. Likewise, adaptive learning rate methods ease this burden by providing automated tuning. AdaGrad~\cite{duchi2011adaptive}, for example, is one of the pioneer adaptive schemes which performs learning rate estimation from the gradients. Other methods are derived from AdaGrad such as AdaDelta~\cite{zeiler2012adadelta}, AdaSecant~\cite{gulcehre2014adasecant}, RMSprop~\cite{tieleman2012lecture} and Adam optimizers~\cite{kingma2014adam}. 
%


Cyclical learning rate (CLR) addresses an issue in training neural networks, i.e., the need to search for the optimal initial rate and subsequent scheduling. The method allows the learning rate to repeatedly swing between boundary limits according to triangle policy that offers more choices in selection of the learning rate. In addition, CLR enhances classification accuracy in a shorter training ~\cite{smith2017cyclical}.

Warmup technique was proposed in early works like ~\cite{vaswani2017attention}, where training utilizes a scheme of starting with a small learning rate and gradually ramping up to a larger value during the number of iterations is much less than the whole length of training ($warmup\_iterations \ll epochs$). The method is built on a theory where the ratio of the learning rate and the batch size affects the dynamics of training. Remark that when training large datasets, we can increase the batch size in order to reduce the training time. However, increasing batch size causes more loss and the benefit of shorter training gained is not proportional to the drawback of increasing loss. For this reason, warmup techniques are relevant in training ConvNets ~\cite{vaswani2017attention,goyal2017accurate,gotmare2018closer,liu2019variance}.
\section{Proposed Methods}
\label{sec:proposed}
\subsection{Collaborative Neural Networks}
\label{sec:collab}
\begin{figure} [htbp!]
\begin{center}
\includegraphics[keepaspectratio,width= 8cm]{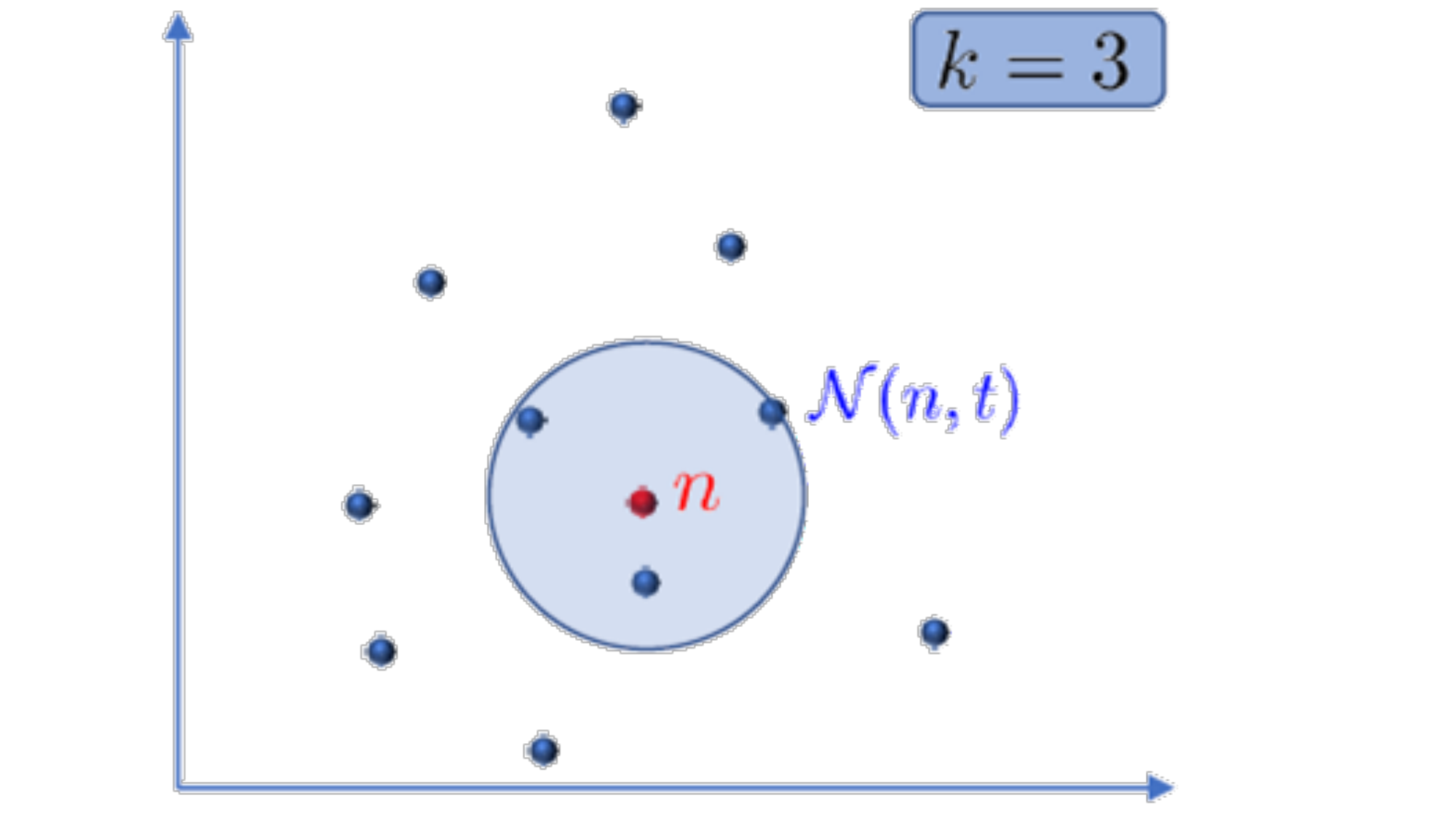}
\caption{Illustration of the three closest particles to particle $n$. The neighborhood $\mathcal{N}(n,t)$ comprises the positions of these particles plus the position of particle $n$ itself.}\label{fig:nn}
\end{center}
\end{figure}

Define $\mathcal{N}(n,t)$ as the set of $k$ nearest neighbor particles of particle $n$ at time $t$, where $k\in\mathbb{N}$ is some predefined number. In particular, $\mathcal{N}(n,t)=\left\{x^{(n)}(t),x^{(i_1)}(t),x^{(i_2)}(t),\ldots,x^{(i_k)}(t)\right\}$, where $i_1$, $i_2$,... $i_k$ are the $k$ closest particles to $n$ and $x^{(i_k)}(t)\in \mathbb{R}^D$ represents the position of particle $i_k$ at time $t$. Figure~\ref{fig:nn} depicts this idea.
Given a (continuous) function $L\,:\,\mathbb{R}^D\longrightarrow \mathbb{R}$ and a (compact) subset $S\subset \mathbb{R}^D$, define
\begin{equation}
\mathcal{Y}={\sf argmin}\left\{L(y)\,:\,y\in S\right\}
\end{equation}
as the subset of points that minimize $L$ in $S$, i.e., $L(z)\leq L(w)$ for any $z\in \mathcal{Y}\subset S$ and $w\in S$.

\begin{figure} [hbt]
\begin{center}
\includegraphics[keepaspectratio,width=8.5cm]{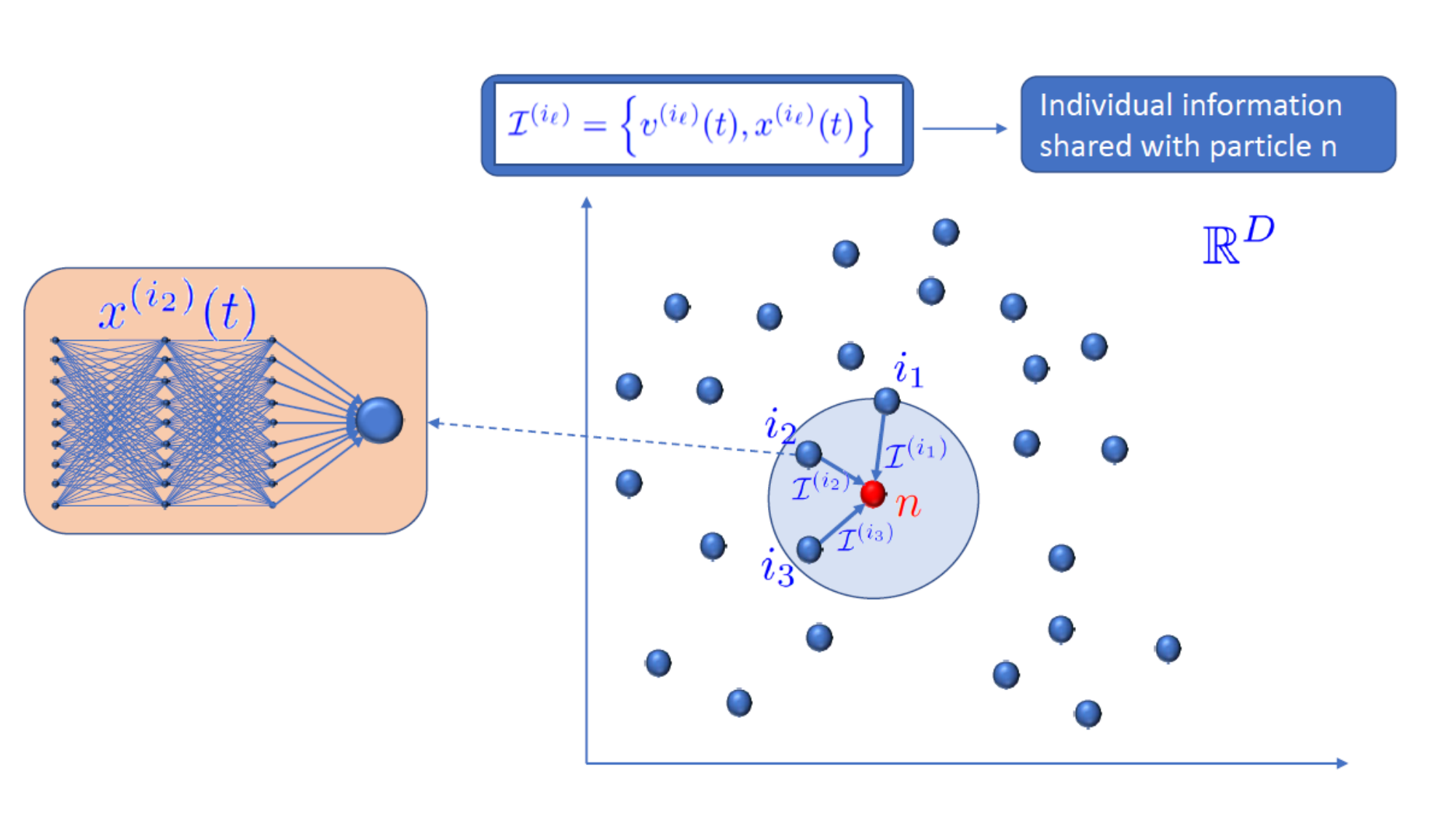}
\caption{Illustration of the PSO phase. At each time instant $t$, particles share information with their current nearest neighbors. In particular, each particle knows at time $t$, the positions and velocities of its neighbors. The position of each particle at time $t$ represents the weights of the underlying neural network.
}\label{fig:nnn}
\end{center}
\end{figure}

We consider a collection of neural networks collaborating in a distributed manner to minimize a Loss function $L$. The neural networks are trained in two phases: i) \textbf{[regular phase]} each neural network is trained via (stochastic) gradient descent; ii)\textbf{[PSO phase]} the algorithm executes an intermediate step of SGD followed by a step of PSO-based cooperation: the vector of weights of each neural network can be cast as the position of a particle in $\mathbb{R}^D$, where $D$ is the number of weights (and dimension of the phase space), and the dynamics of the particles (or neural networks) follow equation~\eqref{eq:PSO}. Figure~\ref{fig:nnn} 
illustrates the general idea. More concretely, the update rule is given by the following dynamics
\begin{equation}
\begin{array}{ccl}
\psi^{(n)}(t+1) & = & -\eta \nabla L\left(x^{(n)}(t)\right)\\ 
& & \\
\phi^{(n)}(t+1) & = & x^{(n)}(t)+\psi^{(n)}(t+1)\\
& & \\
v^{(n)}(t+1) \!\!\! & \!\!\! = \!\!\! & \!\!\! \sum\limits_{\ell \in \mathcal{N}(n,t)} w_{n\ell} \psi^{(\ell)}(t+1) \\
& & \\
& & + c_1 r(t)\left(P^{(n)}(t)-\phi^{(n)}(t+1)\right) \\
& & \\
& & +c_2 r(t)\left(P_g^{(n)}(t)-\phi^{(n)}(t+1)\right)\\
& & \\
x^{(n)}(t+1) & = & x^{(n)}(t)+v^{(n)}(t)
\end{array} 
\label{eq:f1}
\end{equation}
where $v^{(n)}(t)\in\mathbb{R}^{D}$ is the velocity vector of particle $n$ at time $t$; $\psi^{(n)}(t)$ is an intermediate velocity computed from the gradient of the Loss function at $x^{(n)}(t)$; $\phi^{(n)}(t)$ is the intermediate position computed from the intermediate velocity $\psi^{(n)}(t)$; $r(t)\overset{i.i.d.}\sim {\sf Uniform}\left(\left[0,1\right]\right)$ is randomly drawn from the interval $\left[0,1\right]$ and we assume that the sequence $r(0)$, $r(1)$, $r(2)$, $\ldots$ is i.i.d.; $P^{(n)}(t)\in\mathbb{R}^D$ represents the \emph{best} position visited up until time $t$ by particle $n$, i.e., the position with the minimum value of the Loss function over all previous positions $x^{(n)}(0),\,x^{(n)}(1),\,\ldots,\,x^{(n)}(t)$; $P_{g}^{(n)}(t)$ represents its nearest-neighbors' counterpart, i.e., the best position across all previous positions of the particle $n$ jointly with its corresponding nearest-neighbors~$\bigcup_{s\leq t} \mathcal{N}\left(n,s\right)$ up until time $t$: 
\begin{equation}\label{eq:PSO}
\begin{array}{ccl}
P^{(n)}(t+1) & \in & {\sf argmin}\left\{L(y)\,:\,y=P^{(n)}(t),x^{(n)}(t)\right\}  \\
& & \\
P_{g}^{(n)}(t+1) & \in & {\sf argmin}\left\{L(y)\,:\,y=P_{g}^{(n)}(t),x^{(k)}(t);\right. \\
& & \left.k\in \mathcal{N}(n,t)\right\} \\
& &\\
\end{array}.
\end{equation}
The weights $w_{n\ell}$ are defined as
\begin{equation}
w_{n\ell}= f\left(\left|\left|x^{(n)}(t)-x^{(\ell)}(t)\right|\right|\right),
\end{equation}
with $\left|\left|\cdot\right|\right|$ being the Euclidean norm and $f\,:\,\mathbb{R}\rightarrow \mathbb{R}$ being a decreasing (or at least non-increasing) function. We start by assuming that
\begin{equation}
f(z)= \frac{M}{\left(1+z\right)^{\beta}},
\end{equation}
for some constants $M,\beta>0$.

\subsection{Alternative Dynamics}
One alternative to the equation~\eqref{eq:f1} is to \emph{pull back} a particle rather than push the particle in the same direction of gradient. 

\begin{figure} [hbt]
\begin{center}
\includegraphics[keepaspectratio,width=0.45\textwidth]{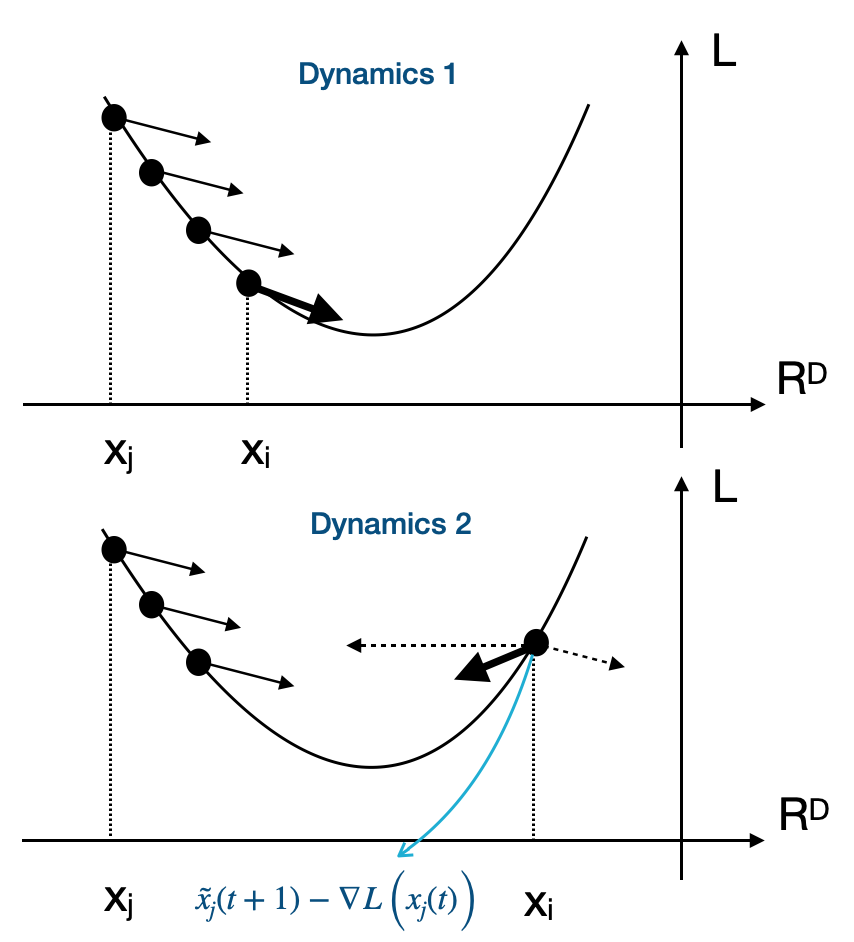}
\caption{Illustration of the Dynamics. The top part of the figure demonstrates the concept of equation~\eqref{eq:f1} whereas the bottom part of the figure shows the concept of equation~\eqref{eq:f2}. The solid line arrows indicate the directions of the particles. The dash line arrows are  vector components.
}\label{fig:dynamics}
\end{center}
\end{figure}

Figure~\ref{fig:dynamics} illustrates the conceptions of the dynamics. In the previous section, particles are assumed to locate in the same side of a valley of the loss function. However, when at least one of the particles lies in an ``antipode'' side of the valley relative to the cluster of remaining particles, the particle is actually pulled away from the minimum by using the first dynamics (Dynamics 1). This is the reason why we introduce the second dynamics (Dynamics 2) to pull the particle back instead. Thus, the formula is as follows:
\begin{equation}
\begin{array}{ccl}
x_{(i)}(t+1) & = & x_{(i)}(t)\\
& & \\
& & + \sum_{j=1}^N \frac{M_{ij}}{(1+\left|\left|x_i(t)-x_j(t)\right|\right|^2)^\beta} (x_j(t)\\
& & - \nabla L(x_j(t)))\\
& & \\
& & + c r\left(P_{nbest(i)}(t)-x_{i}(t)\right) \\
\end{array} 
\label{eq:f2}
\end{equation}
where $x_{(i)}(t)\in\mathbb{R}^{D}$ is the position of particle $i$ at time $t$; $M$, $\beta$ and $c$ are constants decided by experiments with $\left|\left|\cdot\right|\right|$ being the Euclidean norm; $r(t)\overset{i.i.d.}\sim {\sf Uniform}\left(\left[0,1\right]\right)$ is randomly drawn from the interval $\left[0,1\right]$ and we assume that the sequence $r(0)$, $r(1)$, $r(2)$, $\ldots$ is i.i.d.; $P_{nbest(i)}(t)\in\mathbb{R}^D$ represents nearest-neighbors' best , i.e., the best position across all previous positions of the particle $n$ jointly with its corresponding nearest-neighbors~$\bigcup_{s\leq t} \mathcal{N}\left(n,s\right)$ up until time $t$.
\subsection{Further improvements}
As referred before, the vector of weights of each ConvNet plays the role of the position of a particle in phase space. In our framework, we have a \emph{swarm} of ConvNets designed to collaborate in order to attain the best minimum. We discuss further improvements including wider PSO-ConvNet strategy, warmup training and cluster warmup learning rate as follows.
\subsubsection{Wilder PSO-ConvNet Strategy}
\label{sec:randomlrs}
In this section, we propose to improve the dynamics by distinguishing particles based on an essential inner characteristic. Previously, particles are influenced by only distances from their neighbors. However, we argue that some particles might have more effect than the other particles because ConvNets are distinct with intrinsic hyper-parameters (e.g. learning rate, kernel size and number of filters, just to name a few).

Specially, since training ConvNet is an essential phase (intertwining between ConvNet and PSO phases) but has a tendency of getting stuck in local optima, we enable ConvNets to escape the local-minima i.e. a particle accompanied by a learning rate that can freely change (in regular phase). We choose a \textit{random} learning rate over other algorithms as the movement is wildest and much stronger than adaptive learning rates and simpler than the cyclic learning rate. As a result, two types of particles are introduced: i) \textbf{[fast particle]} endowed with a large learning rate; ii) \textbf{[slow particle]} abiding by a small learning rate. A larger learning rate offers a faster training (easier to move out of local minima), but compromises the generalization performance as it incurs greater error. On the other hand, a small learning rate fosters generalization (can be trapped by deeper minima) while requiring a large training time.

Thus, in PSO phase (collaborative phase), a mechanism for conservative particles to be attracted by the wilder particle is proposed. The main idea is simple, by tuning proper weight, the wilder particle would guide other conservative particles to the better minimum.

We restrict the learning rate hyper-parameter to a fixed interval of admissible values. This is motivated by: (i) with a \textit{small} learning rate, the training would take significant longer time; (ii) with a \textit{large} learning rate, the training dynamics becomes unstable compromising convergence to the minima~\cite{hanin2018neural,hochreiter1991untersuchungen}. Therefore, in our formula (equation~\eqref{rnd_generate}), we set the minimum and the maximum values of the learning rate range. Because the learning rate is in an exponential range, e.g., from $10^{-6}$ to $10^{-1}$, we re-scale this range into a linear range and generate learning rates using uniform distribution function so that the small learning rates get an equally likely number of samples as large learning rates. The min and max boundary can be determined via running a learning rate scan between low to high values ~\cite{smith2017cyclical}.
\begin{equation}
\label{rnd_generate}
f(lr_{min},lr_{max})= 10^{U(lr_{min},lr_{max})}
\end{equation}
where $U$ denotes uniform distribution function; $lr_{min}=log_{10}(min)$ and $lr_{max}=log_{10}(max)$.
%
\subsubsection{Warmup Training}
\label{sec:warmuptraining}
Inspired by warmup concept that recently emerged in training deep neural networks~\cite{vaswani2017attention,goyal2017accurate,gotmare2018closer,liu2019variance}, we propose \emph{warump training}. Our idea differs in terms of proportion of warmup time and full training time. In the former, the time required is typically very short, whereas in our technique, warump time often takes a large proportion for ConvNets to train until a desired performance is achieved.

The main reason why we propose warmup training lies in the principle of ConvNets -- training ConvNets takes time. Training on big datasets, e.g., Cifar-10, Cifar-100, SVHN or ImageNet requires from several hours to weeks or months. In addition, we have a few number of ConvNets, so collaboration at an earlier time may reduce opportunity for individual ConvNets to explore more freely. In other words, early collaboration maybe premature. Therefore, we propose to split the warmup training into two steps including (1) a training without intertwining between ConvNet and PSO; (2) and a training with the intertwining. In the second step, training a ConvNet is still needed since the size of ConvNet is huge with millions of weights, thus, a few PSO-ConvNets is certainly not enough for optimization.
\subsubsection{Cluster Warmup Learning Rate}
We attempt to improve the performance of our hybrid PSO-ConvNet by proposing cluster warmup learning rate. Motivated by a conventional training method where we first train a neural network with a large learning rate and gradually reduce the learning rate, in a similar manner, we train all neural networks at a high speed learning rate and then decrease to a slower learning rate. Though, in our approach, the learning rates are obtained from ranges which are generated randomly rather than deterministic.
\section{Experiments}
\label{sec:experiments}
In this section, we will discuss in detail our experiments including chosen benchmark dataset, implementation, how we estimate the learning rate range and evaluation metric.
\subsection{Benchmark Datasets}
\begin{figure}[htb!]
\begin{center}
\includegraphics[width=0.45\textwidth]{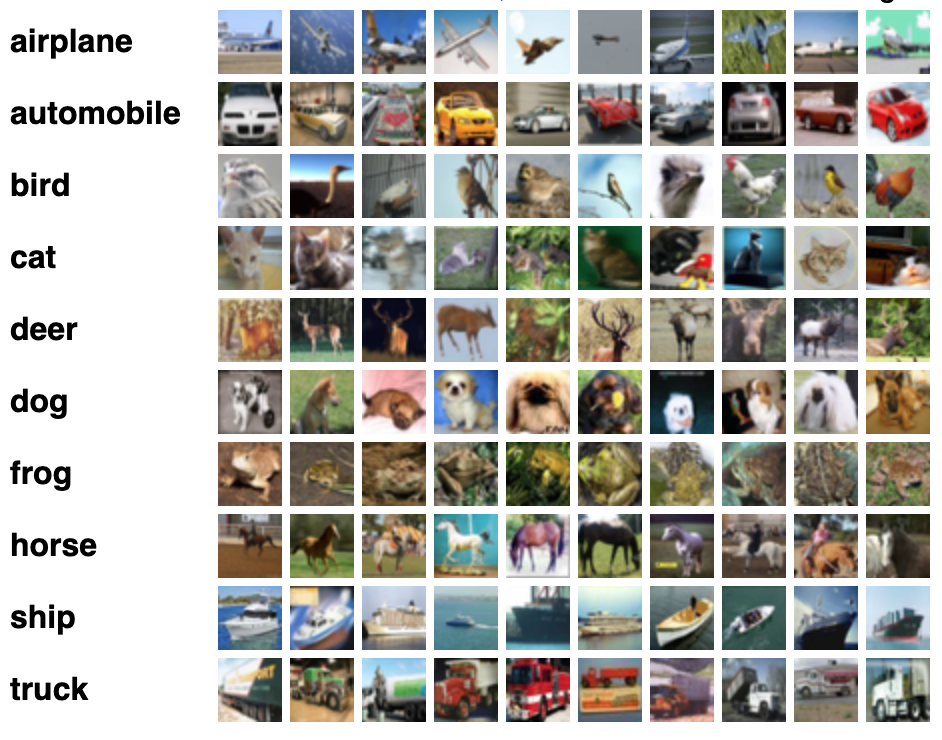}
\caption{A snapshot of samples from Cifar-10 dataset~\cite{krizhevsky2014cifar}.}
\label{fig:cifar10}
\end{center}
\end{figure}

The Cifar-10 and Cifar-100~\cite{krizhevsky2009learning} are popular and challenging datasets for training deep learning. The two datasets contain $50000$ training images and $10000$ testing images with the size of $32 \times 32$ pixels. As the name suggests, Cifar-10 has exactly 10 categories, i.e., airplane, automobile, bird, cat, deer, dog, frog, horse, ship and truck whereas Cifar-100 extends the number of categories from $10$ to $100$. Having the same total number of training images while increasing the number of categories causes Cifar-100 harder to train since there is less images to see in each category than in Cifar-10. Figure~\ref{fig:cifar10} depicts a snapshot of random images from the Cifar-10 dataset.
%
\subsection{Implementation}
In this work, we propose a collaborative training of ConvNets where they share relevant geometric information such as their current gradient and position. In the following subsections, we discuss our proposed Parallel PSO ConvNet approach.
\subsubsection{Parallel PSO ConvNets}
\begin{figure*}[p]
\begin{center}
\includegraphics[keepaspectratio,width=0.9\textwidth]{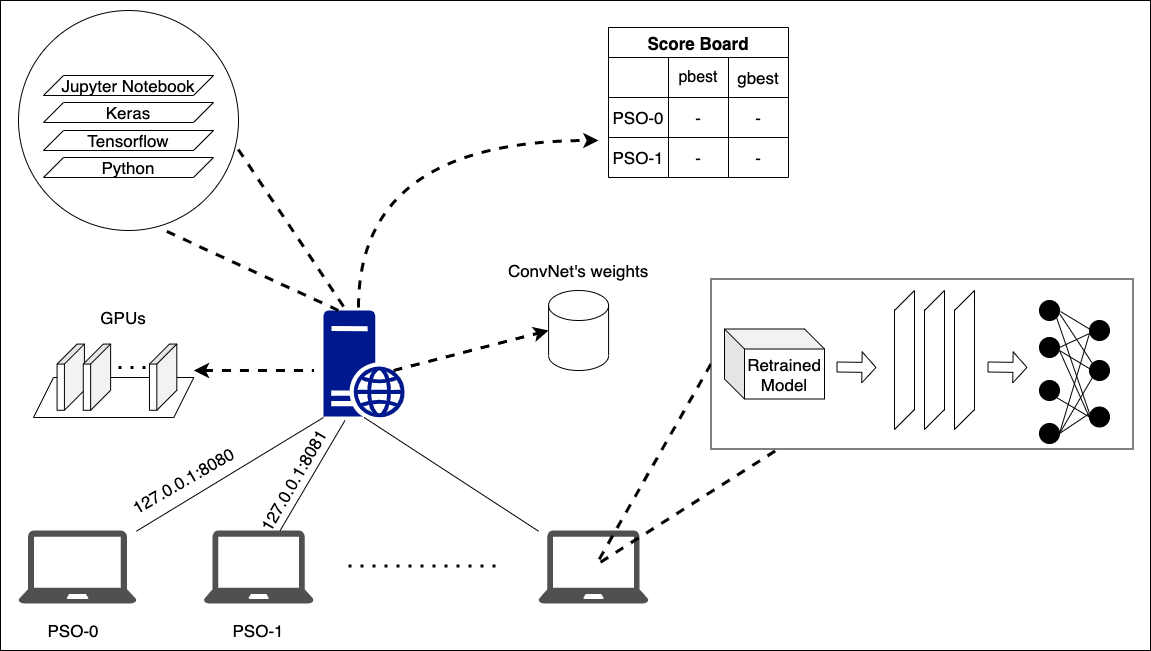}
\caption{Proposed PSO-ConvNets system. PSOs share information with other particles via a file sharing. The information include current location, previous location among the others.}
\label{fig:ecosystem}
\end{center}
\end{figure*}

A crucial aspect of the implementation is to create a distributed environment where particles (ConvNets) cooperate with each other. Figure~\ref{fig:ecosystem} illustrates the design. Typically, ConvNets are often trained using just one local computer or using a remote server. With multiple GPUs, several ConvNets can be trained simultaneously, though, training is performed independently.

We build our PSO-ConvNet design around a web client-server architecture a collaborative training. At the center of the design lies dedicated server which hosts entirely the ecosystem including software stack, modern GPUs (GeForce\textsuperscript{\textregistered} GTX 1080 Ti), network and application layers. Each client connects to one virtual machine in the server via a specific port, and runs a set of procedures so to yield collaborative training of the CNNs which rests primarily on sharing their information: current location, previous location, estimate of the gradient of the loss function, among other elements. As illustrated in the Figure~\ref{fig:ecosystem}, information from each particle are stored in a database, namely ``Score Board''. After each epoch, the latest data will be
inserted into the database for each particle.

More concretely, as entailed in the dynamics of equation~\eqref{eq:f1}, particles will share their current personal best position, the best positions of neighbors as well as their estimate of the gradient. We will further compare our approach with a baseline classical PSO (Section~\ref{sec:psobaseline}) where particles only share their current positions. This will highlight more transparently the importance of sharing the geometric attributes entailed in the dynamics.
\subsubsection{ConvNets}
\label{sec:convnets}
\begin{figure*}[p]
\begin{center}
\includegraphics[keepaspectratio,width=0.9\textwidth]{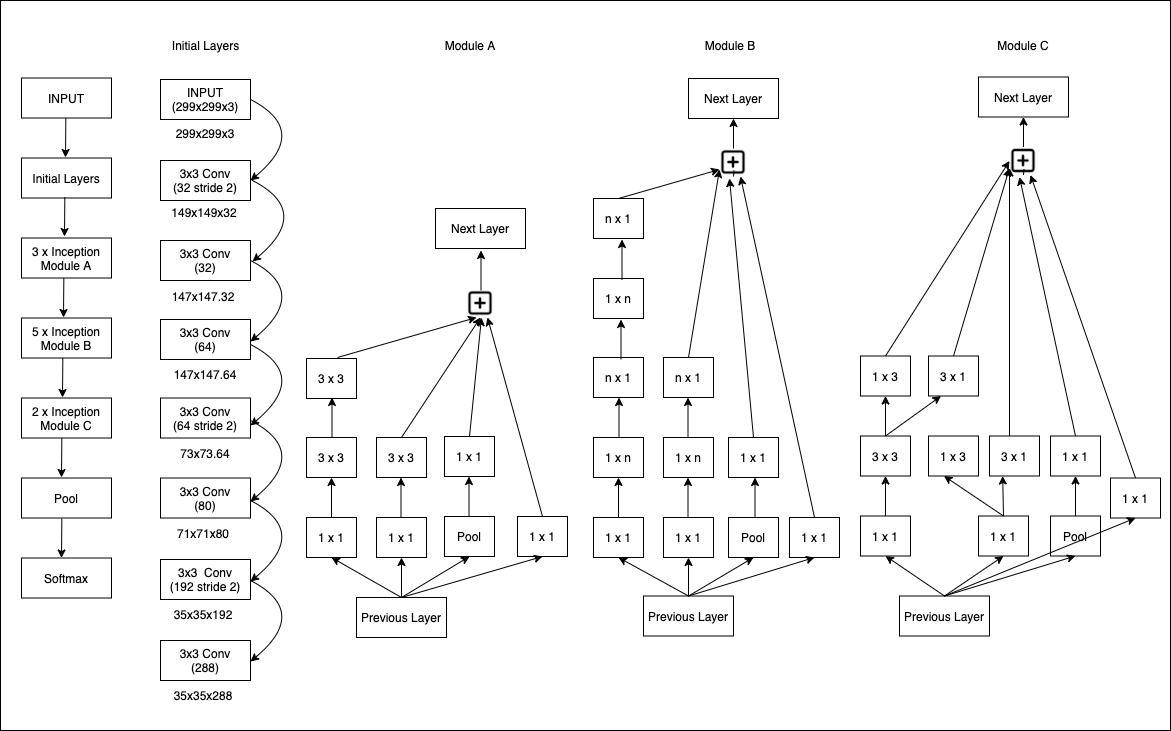}
\caption{Detail of Inception-v3 architecture~\cite{szegedy2017inception} which contains initial layers and several modules A, B and C. Each module comprises factorized convolutions to reduce the computational cost as it decreases the number of parameters.}
\label{fig:inceptionv3_architecture}
\end{center}
\end{figure*}

We consider topologically distinct ConvNets, namely, shallow and deep ConvNets in the mix (Figure~\ref{fig:inceptionv3_architecture} shows architecture of Inception-v3 with 42 layers for illustration purposes). When training ConvNets, transfer learning is often utilized in typical approaches. Though, usually faster, the disadvantage of using transfer learning is that the training stage most likely appears to become stuck in the local minima of the loss function and over-fit the models quickly. Therefore, the optimal procedure here is re-training rather than transfer learning.

Accordingly, all layers of the ConvNets are unfrozen so that the models' architecture and the pre-trained weights (from ImageNet dataset ~\cite{krizhevsky2012imagenet,russakovsky2015imagenet}) can be reused. Following practices from ResNet, Xception and many others, the top layers are replaced and a global pooling is added to reduce the network size. In addition, for enhancing sample variation and also avoiding over-fitting, a small amount of noise is introduced into the networks. 

\begin{figure}[htbp!]
\begin{center}
\includegraphics[keepaspectratio, width=0.45\textwidth]{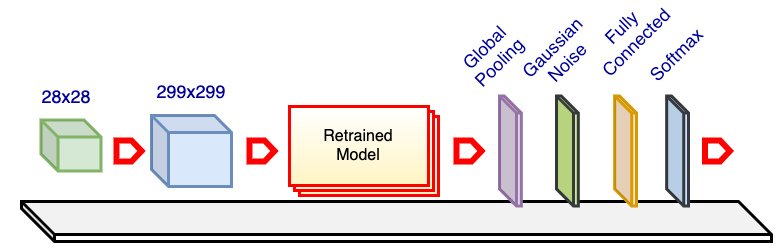}
\caption{Design of re-trained ConvNets architecture. The re-trained model is a pre-trained ConvNet (e.g.: Inception-v3, VGG and ResNet) with the top layers excluded and the weights re-trained. The base model comes pre-loaded with ImageNet weights. The original images are re-scaled as needed to match the required input size of the re-trained model. The global pooling layer reduces the networks size. The Gaussian noise layer improves the variation among samples to prevent over-fitting. The fully connected layer aims to improve classification. The softmax layer is another fully connected layer that has the same number of neurons as the number of dataset categories, and it utilizes softmax activation function.}
\label{fig:retrain_model}
\end{center}
\end{figure}

The re-trained ConvNets architecture is illustrated in Figure~\ref{fig:retrain_model}. The re-trained model is a pre-trained ConvNet (e.g.: Inception-v3, VGG and ResNet) with the top layers removed and the weights re-trained. The base model is a pre-loaded one with ImageNet weights. The original images are re-scaled to match the required input size of the re-trained model. The global pooling layer reduces the network size while the Gaussian noise layer prevents over-fitting. The fully connected layer and softmax layer will improve the classification.
%
\subsection{Estimate learning rates}
\label{experiments0}
For training deep neural networks, the learning rate is one of the most important hyper-parameter as the learning rate decides the amount of gradient to be back propagated or how fast the network moves towards minima. A smaller learning rate requires more training time since only a small change of weights is updated after each epoch whereas a larger learning rate makes training more rapid but also causes more instability. Usually, a shallow method would try out some learning rates and check which one yields a better performance. In contrast, an advanced method would scan from a low to a high value stopping once the Loss increases above a certain threshold (a.k.a learning rate range test)~\cite{smith2017cyclical}.

\begin{figure}[htbp!]
\begin{center}
\includegraphics[keepaspectratio,width=0.4\textwidth]{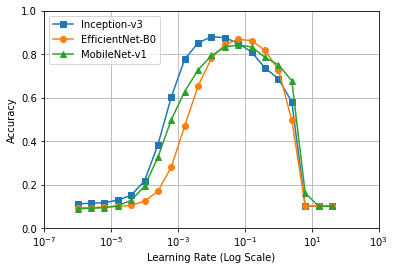}
\caption{Learning rate range scan for Inception-v3, EfficientNet-B0 and MobileNet-v1 on Cifar-10 dataset.}
\label{fig:lr_scan}
\end{center}
\end{figure}

Figure ~\ref{fig:lr_scan} shows results of this learning range test using three different ConvNets namely Inception-v3, EfficientNet-B0 and MobileNet-v1 on Cifar-10 dataset. 
%
Though slightly different, we can glean that accuracy drastically increases and decreases at around [$10^{-5}$, $10^{-4}$] and [$10^{-2}$, $10^{-1}$], respectively. Therefore, we set $10^{-5}$ and $10^{-1}$ as the minimum and the maximum boundaries for PSOs with random learning rates.
\subsection{Evaluation Metric}
In this paper, we evaluate the visual classification performance of different algorithms using overall accuracy. The metric is popular for the comparison and analysis of results in computer vision tasks. The metric is defined as follows:
\begin{equation}
Accuracy=\frac{\mbox{Number of correct predictions}}{\mbox{Total numbers of predictions made}}.
\end{equation}
%
%
\section{Classification results}
\label{sec:results}
In this section, we discuss the results of our proposed hybrid PSO-ConvNet according to equation~\eqref{eq:f1} which we call Dynamics~$1$ and also the approach with the interplay of SGD and PSO according to equation~\eqref{eq:f2} which we call Dynamics~$2$. We refer to as the terms ``Dynamics'' to emphasize an essential property of the formulas where a particle adjusts its direction towards the average of its neighbors' directions. It means that the particle weighs more nearer particles than those farther away.
\subsection{Dynamics~$1$}
In this subsection, we discuss the results of our proposed hybrid PSO-ConvNet according to equation~\eqref{eq:f1}. For easiness of reference to each element in that equation, we identify the first element as \textbf{Gradient} since the element mainly relates to vectors of particles; the second element as \textbf{Personal best (pBest)} and the last element as \textbf{Nearest Neighbor best (KNN)}, just because the elements describe the best locations obtained by individual particles and group of nearby neighbor particles.

We first see the results with the influence from the nearest neighbors, then from the gradients and at last the combination of the two. Further, we will enable the pBest term to see the effects.  In the experiments with KNN, we keep the accelerator $c_2$ as a constant $0.5$ and vary the number of nearest neighbors from $1$ to $3$. In case we have only one nearest neighbor it means that a particle shares information with $1$ neighbor ($k=1$). The same also applies for other numbers of neighbors. In the experiments with Gradient, we select $M$ as $0.1$, $1$ and $10$ while $\beta$ is set to $1$. 
The results with the above settings are summarized in Table~\ref{tab:table_basic_experiments}. Interestingly, it is easy to observe that a combination of KNN and Gradient generates a higher accuracy (approximately $0.9800$) than using either KNN or Gradient alone ($0.9780$).

\input{table_initial_experiments}

\begin{figure*}[htb!]
\begin{center}
\includegraphics[keepaspectratio,width=0.85\textwidth]{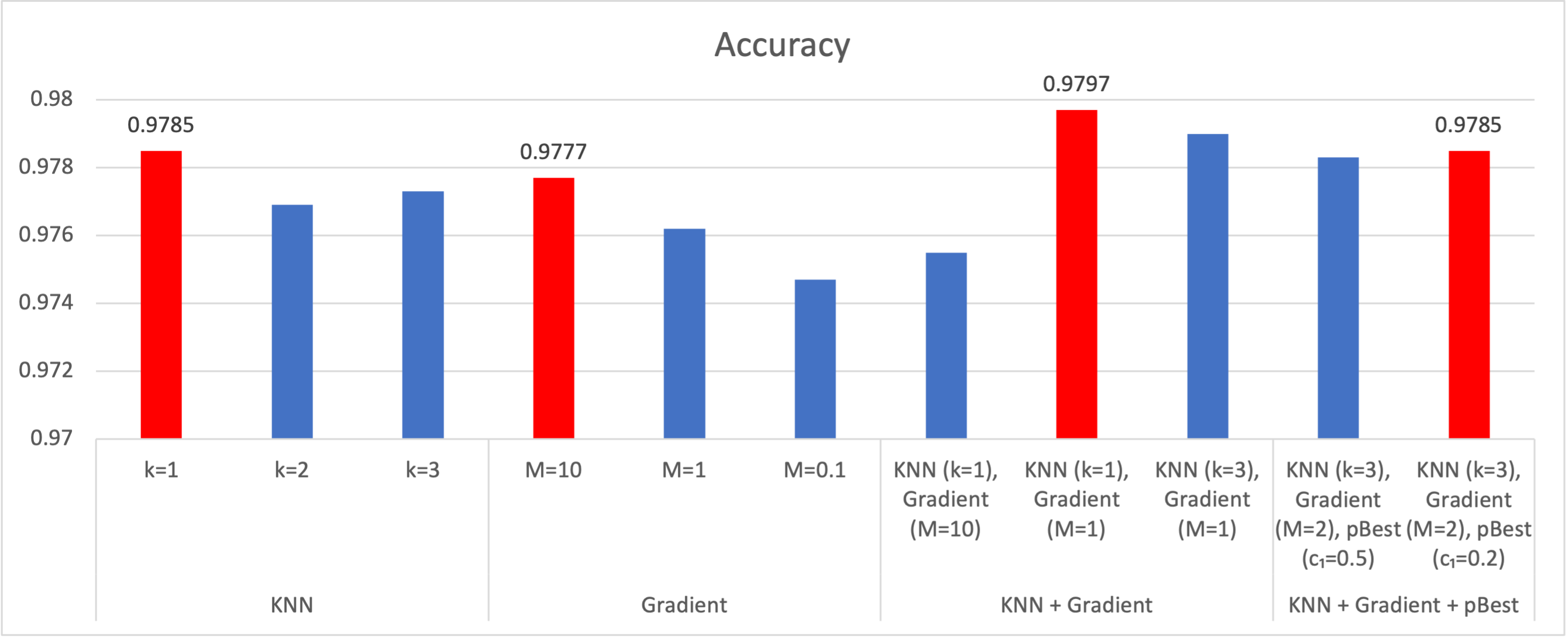}
\caption{Proposed PSO-ConvNets Dynamics~$1$ (equation~\ref{eq:f1}). Comparison when KNN and Gradient are tested separated and combined, and also with the inclusion of pBest. Each column summaries the best accuracy of all PSOs (PSO-1, PSO-2, etc.). The red column highlights the best result for each group (KNN, Gradient, etc.).}
\label{fig:knn_gradient_pbest}
\end{center}
\end{figure*}

In Figure~\ref{fig:knn_gradient_pbest}, we graphically show the proposed PSO-ConvNets with Dynamics~$1$. Similarly, as referred above, the results show higher accuracy with the combination of KNN and Gradient. However, with the inclusion of pBest, it is observed that the accuracy overall decreased. Since adding more elements would cause our experiments to grow exponentially, we decided to discard the pBest element. 
Thus, from here onwards, the Dynamics~$1$ will include only Gradient and KNN elements and exclude pBest unless stated otherwise.
%
%
%
%
\begin{figure}[htb!]
\begin{center}
\includegraphics[keepaspectratio,width=0.39\textwidth]{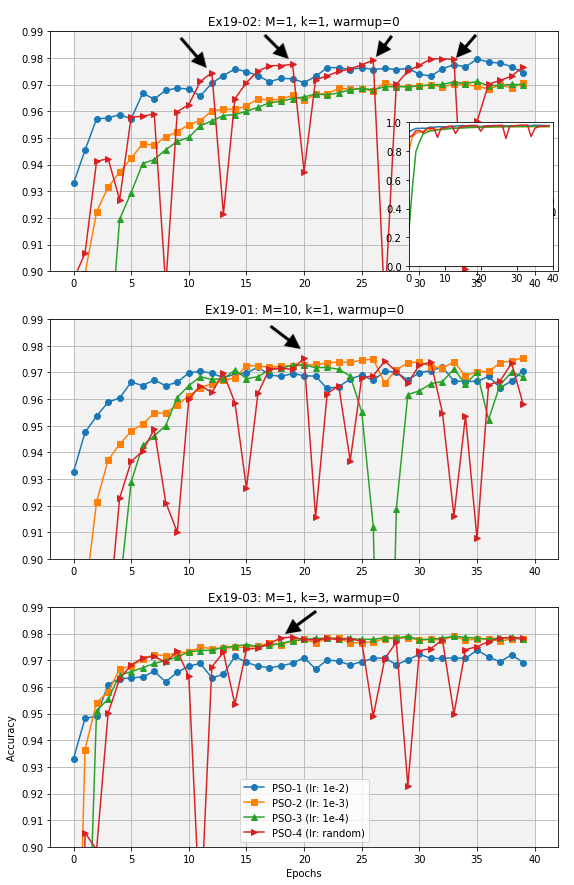}
\caption{Accuracy during training of Hybrid PSO-ConvNets. Examples when $M=1$ and $M=10$, ${\sf warmup}=0$ and on variation of k ($k=1$ and $k=3$). Learning rates of PSO-1, PSO-2 and PSO-3 are set at $10^{-2}$, $10^{-3}$ and $10^{-4}$ while PSO-4's learning rate is random in the range [$10^{-5}$, $10^{-1}$]. The arrows indicate where PSO-4 finds better locations. The zoom in figure displays the accuracy in full scale whereas the others show a narrower scale for better view in detail.}
\label{fig:accuracy}
\end{center}
\end{figure}

In Figure~\ref{fig:accuracy}, we demonstrates the advantage of random learning rate as at several times the PSO achieves higher accuracy than the nearest neighbor (see the indicated arrows' location). For example, in case of PSO-1 and PSO-4 couple in the upper part of the figure, the PSO-1 appears to be stuck at a local minimum between epoch 5th and 10th, then PSO-4 finds a better accuracy.
This occurrence repeats several times during the training even though PSO-4 goes to worse accuracy areas. 
\subsection{Optimization of parameter $M$}
\label{sec:M}
We evaluate the performance of the hybrid PSO-ConNets based on the variation of parameter $M$. In our experiment, rather than fixing a value $M$ for every particle as in \cite{cucker2007emergent}, the exchanged gradients are properly weighted. Thus, we further sharpen the capability of the algorithm which is not only distinguishing particles based on distance but also taking into account other intrinsic attributes of the particles. In this sense, a particle can be incorporated with a large random learning rate or sometimes called \textit{wilder} particle since the particle can scan the landscape faster and also goes deeper in local minima. For the capability, this particle should be attracted more by other \textit{conservative} particles. Or in other words, particles would be pulled faster toward the gradient directions of the wilder PSOs than to other particles.

Initially, we set the weight $M$ between conservative particles being small numbers (e.g.: $0$ and $0.2$); in the case of conservative and wilder (with more liberty) particles or among wilder particles we set larger values for $M$ (e.g.: $1.2$ and $1.7$). 

\begin{figure}[tb!]
\begin{center}
\includegraphics[keepaspectratio,width=0.45\textwidth]{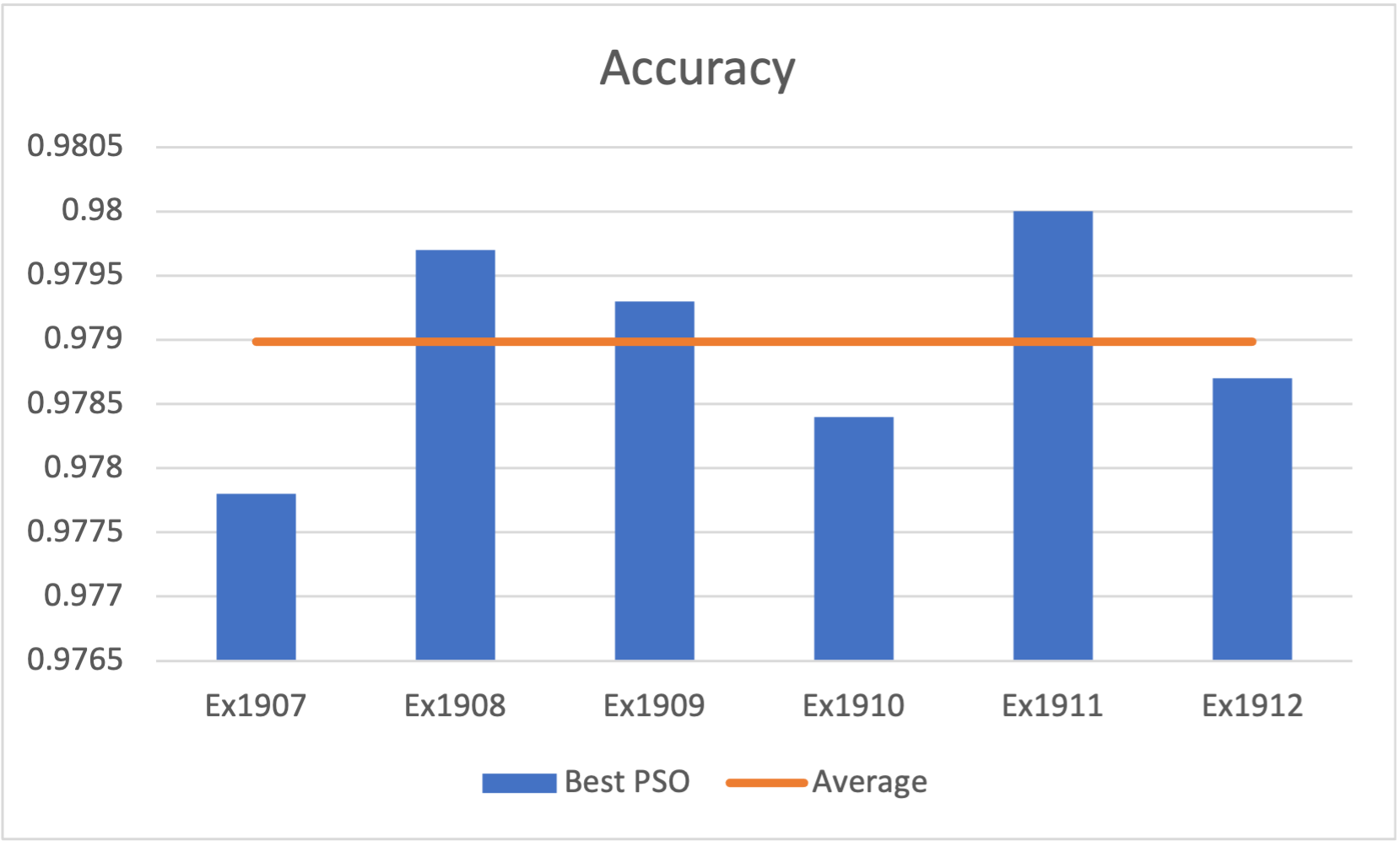}
\caption{Results when $c_2$ and $warmup$ are fixed at $0.5$ and $0$, respectively.}
\label{fig:m1}
\end{center}
\end{figure}

Figure~\ref{fig:m1} shows the results when $c_2$ and $warmup$ are fixed at $0.5$ and $0$. We recall that accelerator $c_2$ controls the effect of KNN element in the equation~\eqref{eq:f1}, i.e., the faster $c_2$ the more significant the element is. In the meantime, $warmup$ indicates at which epoch all particles begin to collaborate. For each experiment, we summary the highest accuracy among PSOs as "Best PSO". Further, we notice that on average the accuracy settles at approximately $0.9790$ and the best accuracy rises up to $0.9800$.

\input{table_adv_experiments}

\begin{figure*}[htb!]
\begin{center}
\includegraphics[keepaspectratio,width=0.75\textwidth]{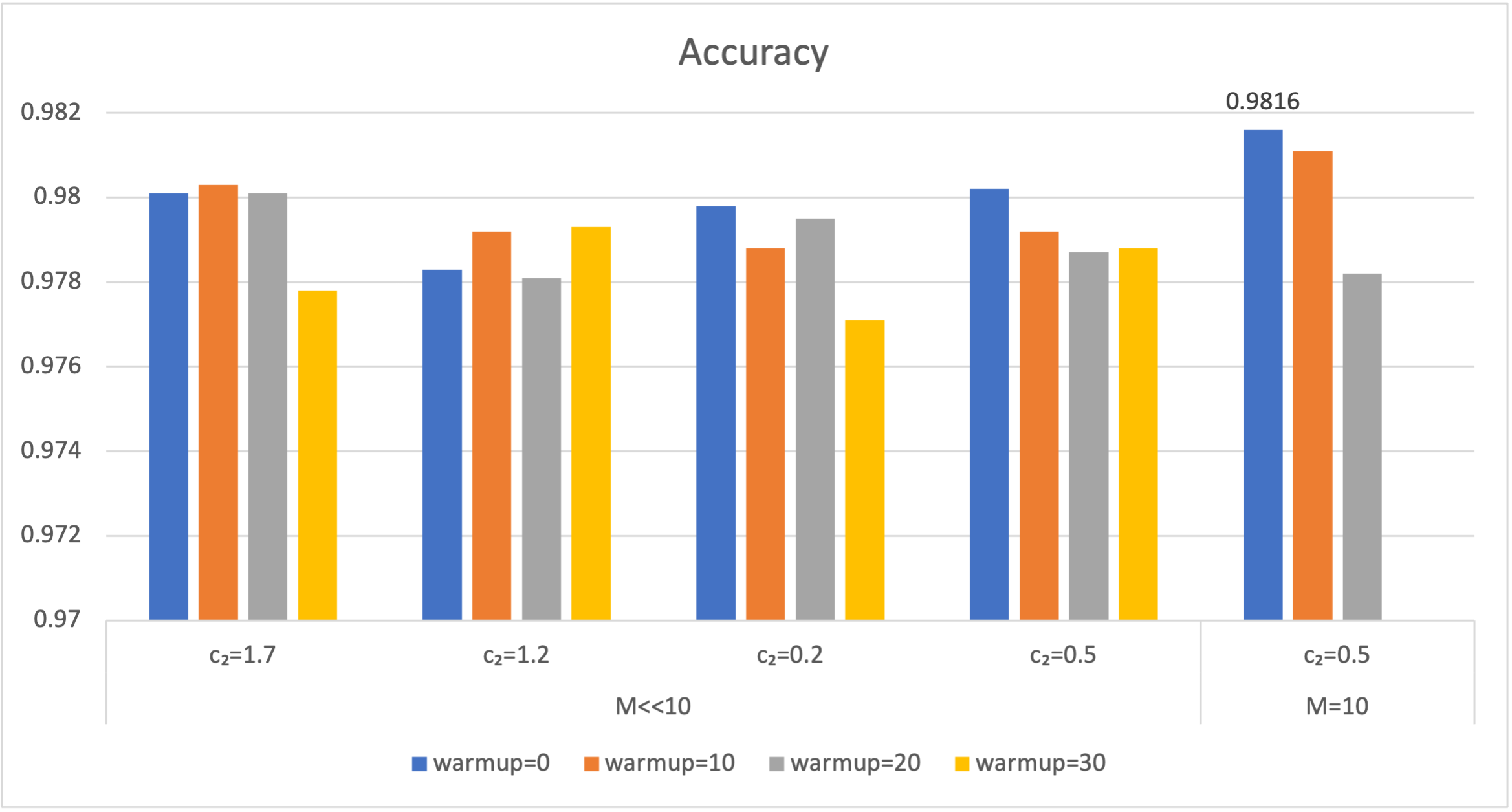}
\caption{Results on variation of $c_2$ and $warmup$. The weight $M$ from all particles toward wilder particles is set at a higher value. The best accuracy is recorded for all PSOs in each experiment.}
\label{fig:m2}
\end{center}
\end{figure*}

We then increase the weight towards wilder particles to a much higher value ($M=10$). The results for variety of $c_2$ and $warmup$ are shown in Table~\ref{tab:gradients1} and also plotted on Figure~\ref{fig:m2} to facilitate the analysis. On the left-hand side, among variations, $c_2$ equals to $1.7$ and $0.5$ accomplish better accuracy than that of $1.2$ and $0.2$. In addition, pulling particles with distinct weights favors those warmups that start at the beginning of training (at $c_2=0.2$ and $c_2=0.5$). Remarkably, on the right-hand side where $M$ is greatly increased, we achieve a milestone result with accuracy of $0.9816$.

\begin{figure}[htb!]
\begin{center}
\includegraphics[width=0.5\textwidth]{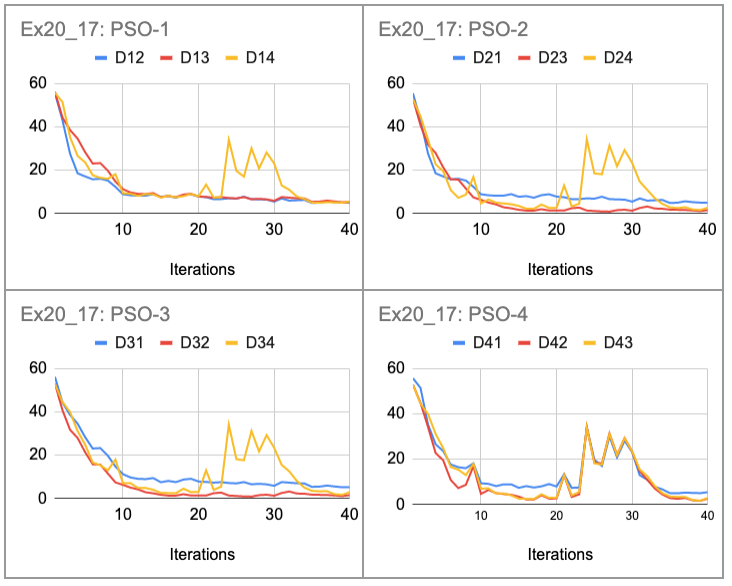}
\caption{Distances between PSOs. $D_{ij}$ denotes the distance between particle $i$ and particle $j$. Learning rates of PSO-1, PSO-2 and PSO-3 are set at $10^{-2}$, $10^{-3}$ and $10^{-4}$ while PSO-4's learning rate is random in the range [$10^{-5}$, $10^{-1}$].}
\label{fig:distance}
\end{center}
\end{figure}

Next, we look at the dynamics of change in distances between PSOs for the best result above as shown in Figure~\ref{fig:distance}. We can see that PSO-2, PSO-3 and PSO-4 gradually approach each other (the distances to PSO-1 have similar values). In addition, PSO-1, though also approaches the group, keeps a further distance. It indicates that possessing a larger learning rate ($10^{-2}$ vs. $10^{-3}$ and $10^{-4}$) prevents the PSO from going into steeper minima. Additionally, PSO-4 (random PSO) scans the solution space more broadly as the distances fluctuate wilder, even sometimes move out of the group.
\subsection{Optimization of parameter $\beta$}
\label{sec:beta}
In this experiment, we attempt to improve the performance of our proposed PSO-ConvNets by changing the parameter $\beta$ in Dynamics~$1$. The parameter regulates the rate of decay (in the gradient) which is a distinct feature in comparison with kinetics gradients. In addition, $\beta$ amplifies the effect of the distance, i.e., setting a smaller value will have a similar effect as when the particles stick together whereas a bigger value will separate the particles.

\begin{figure}[htb!]
\begin{center}
\includegraphics[keepaspectratio,width=0.45\textwidth]{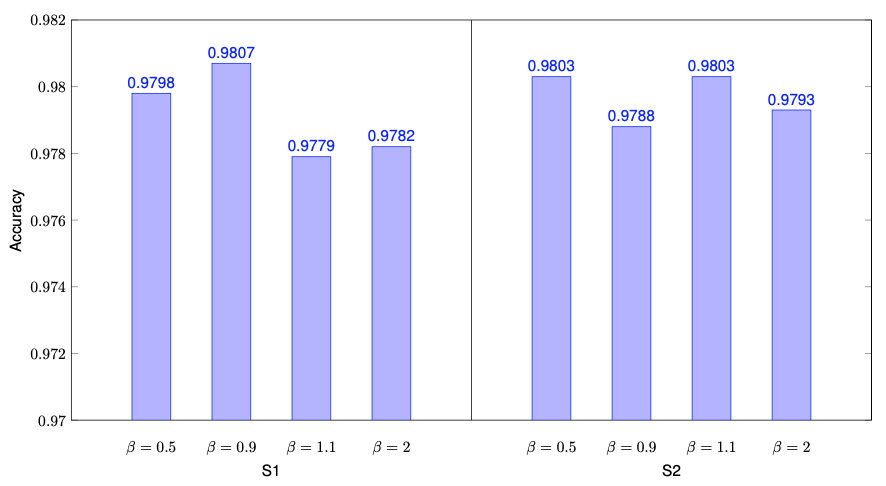}
\caption{Analysis of $\beta$. The best accuracy for all PSOs are recorded for each experiment. Besides, $k$, $c_2$ and $warmup$ are set at 3, 0.5 and 0, respectively.}
\label{fig:beta}
\end{center}
\end{figure}

\begin{figure*}[htb!]
\begin{center}
\includegraphics[keepaspectratio,width=0.95\textwidth]{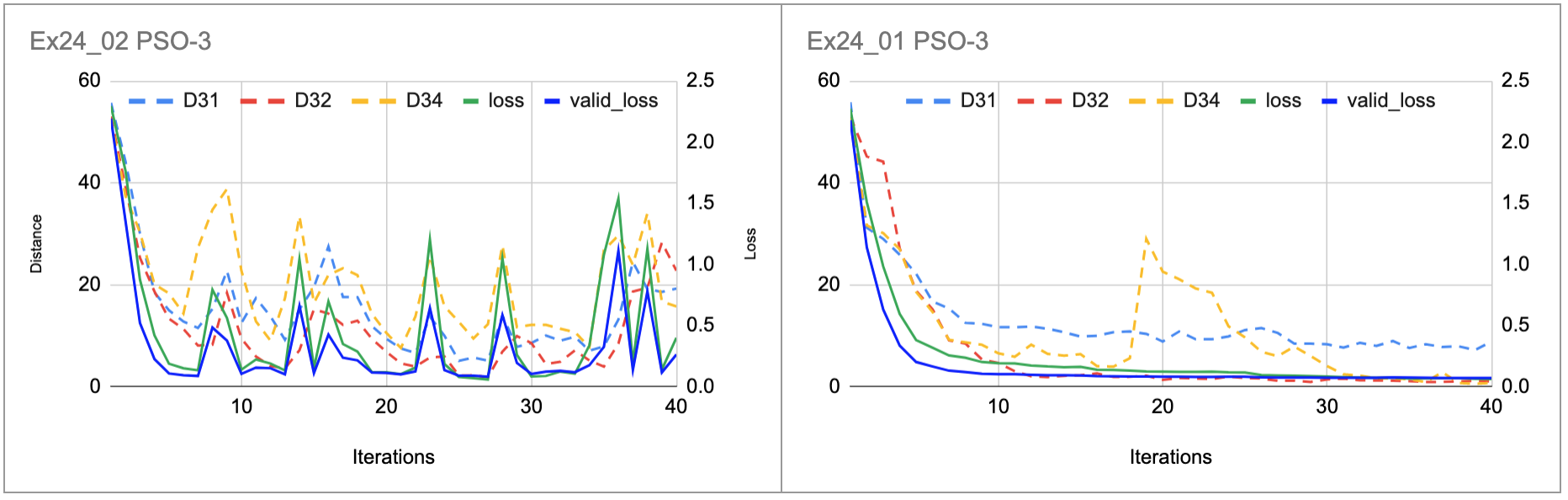}
\caption{Effects of $\beta$ to distances between PSOs for $\beta=0.5$ and $\beta=2$.}
\label{fig:loss}
\end{center}
\end{figure*}

Figure~\ref{fig:beta} shows results using different $\beta$ values ($0.5$, $0.9$, $1.1$ and $2$). As an aside remark, two experimental settings,  S1 and S2, are described in Table~\ref{tab:gradients}. Though, the accuracy is not better than the one found in the previous section, we notice an important qualitative behavior when we look in more detail which is illustrated by inspecting  both graphs in Figure~\ref{fig:loss}. In fact, we select particle PSO-3 for illustration since its dynamics changes more than the others. In the left graph ($\beta=0.5$), it appears to have more disturbances than on the right one when $\beta=2$. At points where PSO-3's loss increases, the particle moves away from the group as the distances from the particle to all the others increase. The behavior does not appear on the right graph. It seems that when particles move closer, a particle is pushed away in the opposite direction. This is the reason why we propose the second dynamics (equation~\eqref{eq:f2}) whose results will be discussed in Section~\ref{sec:d2}.
\input{table_settings}
\subsection{Number of nearest neighbors}
\begin{figure*}[htb!]
\begin{center}
\includegraphics[keepaspectratio,width=0.95\textwidth]{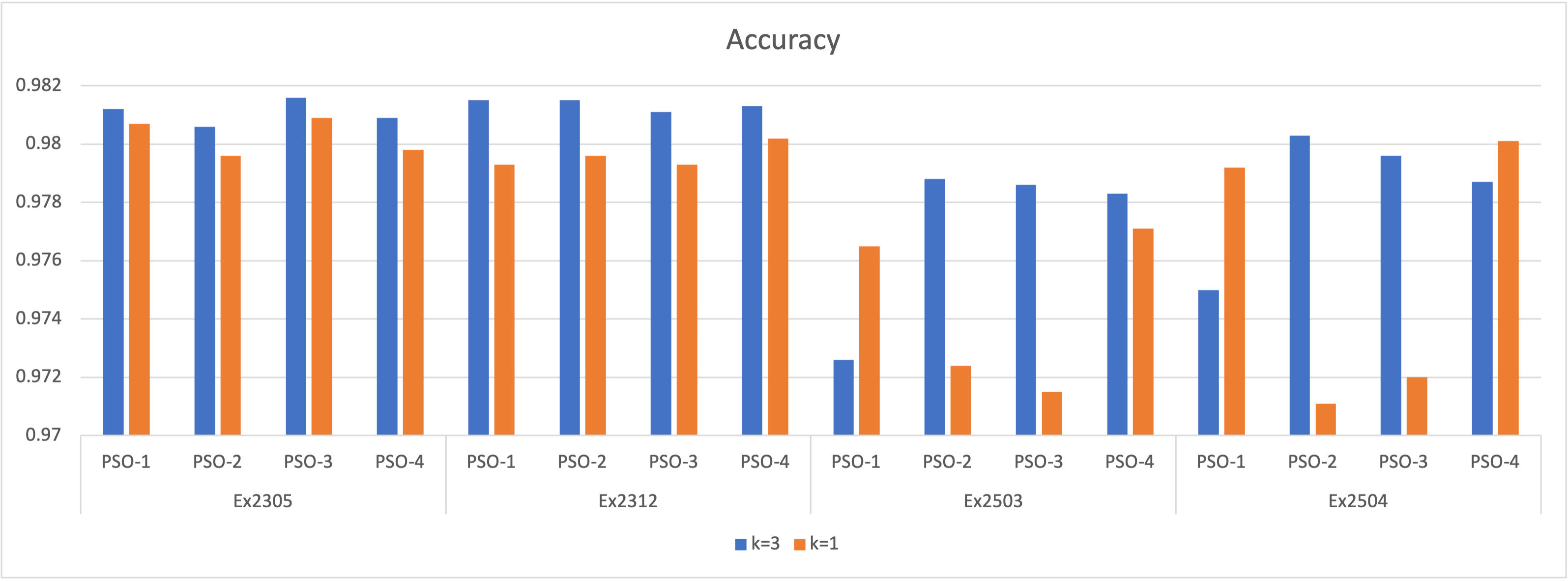}
\caption{Comparison accuracy performance when number of nearest neighbors $k=3$ and $k=1$.}
\label{fig:k}
\end{center}
\end{figure*}

To study the performance of PSO-ConvNets with respect to the number $k$ of nearest neighbors,  $k=1$ and $k=3$  are applied (we exclude $k=2$ because in this case, the random PSO will be kept far from the group of the other three and we need the PSO to attract other PSOs). When $k=1$, a particle obtains information from only one nearest neighbor while when $k=3$, all the particles cooperate with each other (we recall that we assume a total of four particles in this work). The classification accuracy is illustrated in Figure~\ref{fig:k}. We see that the accuracy is overall higher when $k=3$, i.e., when all particles exchange information among all the other particles. However, we assume that if the number of PSOs could be scaled up to a much larger number, obtaining data from all neighbors would be more expensive than from just some nearest neighbors. Therefore, in the large-scale setting, meeting a trade off between the number of neighbors and efficiency would be a better strategy.
\subsection{Neural Networks' baseline}
\label{neuralnetworks_baseline}
Our proposed PSO-ConvNet approach is comprised of two phases in which the first phase refers to the training of each ConvNet independently and the second phase refers to the collective training of the ConvNets wherein each ConvNet integrates information exchanged by neighboring ConvNets following a PSO inspired approach. In this experiment, we focus on analysis of training ConvNets and leave out PSO phase. As we described in Section~\ref{sec:convnets}, a ConvNets is built on a re-trained model instead of transfer learning model because the training is faced with over-fitting after a short time. In addition, we unfrozen the layers of ConvNets so that we can re-use the model's architecture and weights. Retraining a model from scratch would take more time than re-using weights. In this sense, we compare the performance of a re-trained model (model with unfrozen layers) to a transfer learning model (model with frozen layers). We also call the transfer learning model as a baseline. Furthermore, to enable augmentation for image input, we train the baseline in only one step. This means that an input after feature extraction is continuously becoming the input for the next layers (usually comprises of global pooling and fully connected layers). In transfer learning, training usually is separated into two separated steps which includes feature extraction and fine-tuning.
\input{table_parameters}
\begin{figure*}[htb!]
\centering
\includegraphics[keepaspectratio,width=0.90\textwidth]{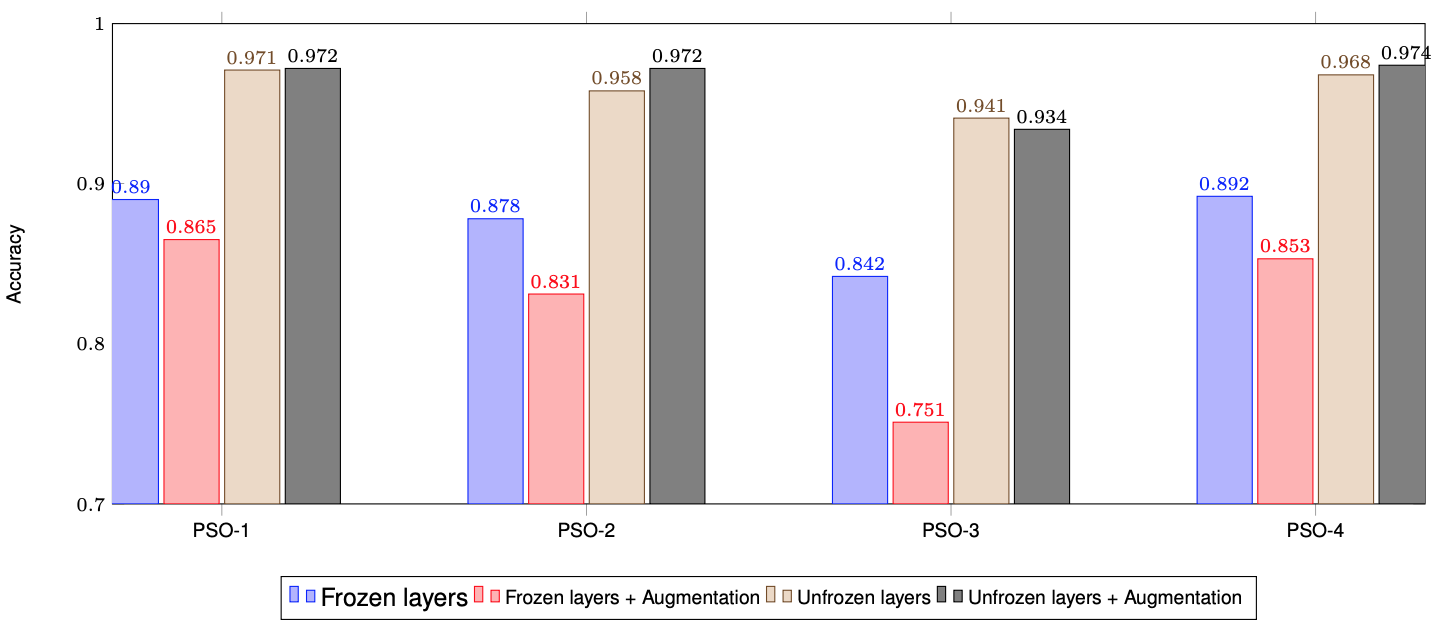}
\caption{Results of ConvNets when the main layers of networks are frozen and unfrozen along with augmentation. Learning rates of PSO-1, PSO-2 and PSO-3 are set at $10^{-2}$, $10^{-3}$ and $10^{-4}$ while PSO-4's learning rate is random in the range [$10^{-5}$, $10^{-1}$]. ConvNets are not collaborated in this experiment.}
\label{fig:baseline_neural_networks}
\end{figure*}
\begin{figure*}[htb!]
\begin{center}
\includegraphics[keepaspectratio, width=0.85\textwidth]{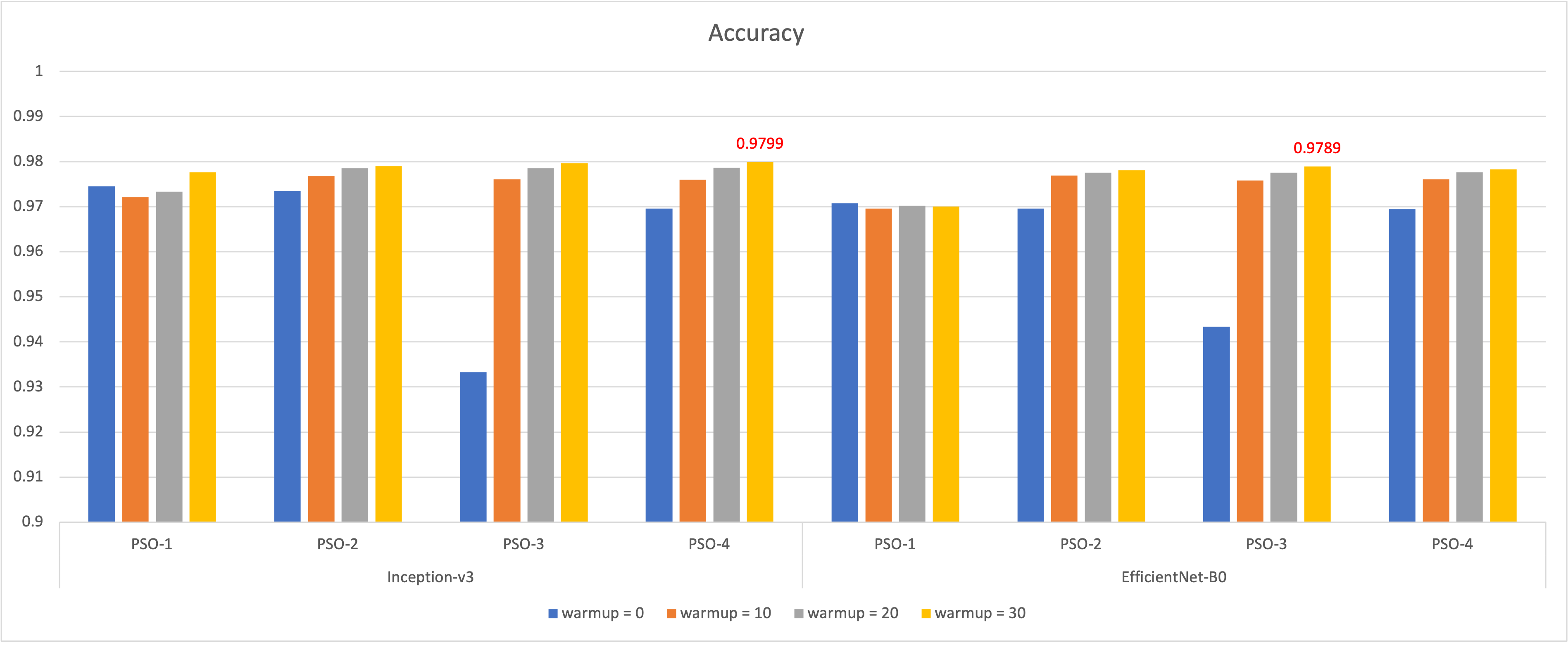}
\caption{PSO gBest. Performance of Inception-v3 and EfficientNet-B0 on several values of $warmup$.}
\label{fig:baseline_pso}
\end{center}
\end{figure*}
The parameters for this experiment are described in Table~\ref{tab:table_parameters} and can be categorized into two groups, namely, ConvNets and Augmentation. The first one concerns internal settings for ConvNets including the lengths of training, Gaussian noise level and size of fully connected layer as well as batch size. We set the number of iterations at 40 since the training appears to start over-fitting by that iteration. We also add Gaussian noise as well as carefully choosing the number of neurons in fully connected layer to reduce over-fitting. The batch size is set at 32 to take full capability of GPUs' memory. The second group controls the augmentation for ConvNets including standard techniques such as rotation range, width shift range, height shift range, shear range and zoom range. Besides, channel shift range, fill mode and preprocessing also makes ConvNets more robust.

Since a standard for Cifar-10 architecture and hyper-parameter settings are not available in Keras ecosystem (a famous API to develop deep learning applications) to the best of our knowledge, we decide to build our models based on popular Inception-v3 structure and compare our results with recent state-of-the-art using the same architecture. 

First, as we can glean from Figure~\ref{fig:baseline_neural_networks}, the results of transfer learning models using frozen layers are approximately $0.89$. In comparison with recent works, the authors in~\cite{alkhouly2021improving} and~\cite{kalayeh2019training} reported accuracy of $0.86$ and $0.92$, respectively. Thus, the differences are in a decent range and we will use the model as a baseline to improve our work. Second, we see that results of unfrozen neural networks (re-trained model) outperform those with frozen layers by a large margin increasing by roughly $8\%$ in settings with and without augmentations.
\subsection{PSO gBest}
\label{sec:psobaseline}
We next test a PSO using global best approach (namely, PSO gBest) similar to Dynamics~$1$. While the training is still intertwined between PSO and SGD but the differences are that (i) the gradient element is excluded and (ii) instead of updating particle's location toward a nearest best neighbor's location, particles in gBest will be updated toward the global best as in the following formula.
\begin{equation}
v^{(n)}(t+1) = v^{(n)}(t)+c r(t)\left(P_{gBest}^{(n)}(t)-\phi^{(n)}(t+1)\right)
\label{eq:gbest}
\end{equation}
where $v^{(n)}(t)\in\mathbb{R}^{D}$ is the velocity vector of particle $n$ at time $t$; $r(t)\overset{i.i.d.}\sim {\sf Uniform}\left(\left[0,1\right]\right)$ is randomly drawn from the interval $\left[0,1\right]$ and we assume that the sequence $r(0)$, $r(1)$, $r(2)$, $\ldots$ is i.i.d.; $P_{gBest}^{(n)}(t)$ represents its global best, i.e., the best position across all previous positions of all particles up until time $t$.

In these experiments, we set $c$ as a constant with a value of $0.5$. Besides, ConvNets cooperate from the beginning and at iterations 10th, 20th and 30th. We also utilize two distinct ConvNets' architectures, i.e., Inception-v3 and EfficientNet-B0 for comparison. The results are shown in Figure ~\ref{fig:baseline_pso}. Overall, Inception-v3 performs slightly better than its counterpart, e.g., particle PSO-4 in Inception-v3 structure achieves an accuracy of approximately $0.9799$ when $warmup=30$ while the latter obtains a smaller value ($0.9789$). It is also interesting to notice that delaying cooperation at a later time mostly brings accuracy to a higher value. This differs from the results in Section~\ref{sec:M} where collaborating earlier is generally better. One explanation is that, in Dynamics~$1$, a particle is attracted due to more causal effects, e.g., direction to the best location, directions of other particles so the particle spans more landscape in finding solutions, thus, training earlier is essential.
\subsection{Accelerator coefficient}
\begin{figure}[hbt!]
\begin{center}
\includegraphics[keepaspectratio,width=0.45\textwidth]{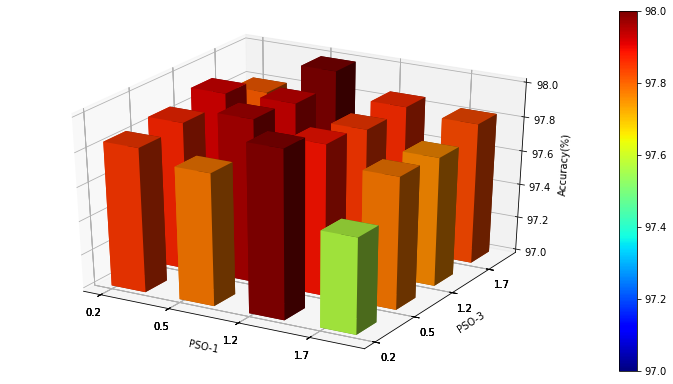}
\caption{Evaluation of PSO's accelerator. The results are obtained from a re-trained EfficientNet-B0 model using gBest approach. The accelerators of PSO-2 and PSO-4 are fixed at $0.5$ while PSO-1 and PSO-3 vary.}
\label{fig:gbest}
\end{center}
\end{figure}

We try to find a connection between accelerators $c$ among PSOs and report accuracy of PSO gBest method in Figure~\ref{fig:gbest}. Due to exponential growth in number of experiments, we evaluate only two PSOs rather than using all four PSOs. We choose PSO-1 and PSO-3 since these PSOs are set at fastest and slowest learning rates (PSO-4 is excluded because of instability). We also select the ConvNets' architecture EfficientNet-B0 because its network size is smaller, i.e., each neural network has about five million total number of parameters in comparison with twenty four millions that of Inception-v3. According to the learning rate range scan in Figure~\ref{fig:lr_scan}, the speed needs to be faster. Thus, we change  PSO-1, PSO-2 and PSO-3's learning rates to $10^{-1}$, $10^{-2}$ and $10^{-3}$, respectively. In the same manner, the range for PSO-4 is also moved to between $10^{-1}$ and $10^{-5}$. We can observe that, for certain settings, e.g., when PSO-1's accelerator equals to $1.7$, the overall accuracy seems to be reduced and when the value is $0.5$, the accuracy appears to be increased.
%
\subsection{Additional strategies for improvement}
\subsubsection{Multi-wilder particles}
Now, we consider particles with larger random learning rate (\emph{wilder} particles) and particles with fixed learning rates (\emph{conservative} particles). In this experiment, the number of wilder particles expands to more than just one. Thus, we expect that more wilder particles would scan more the solution space, consequently find better minima and to improve overall performance.
%

\begin{figure*}[htb!]
\begin{center}
\includegraphics[keepaspectratio,width=0.85\textwidth]{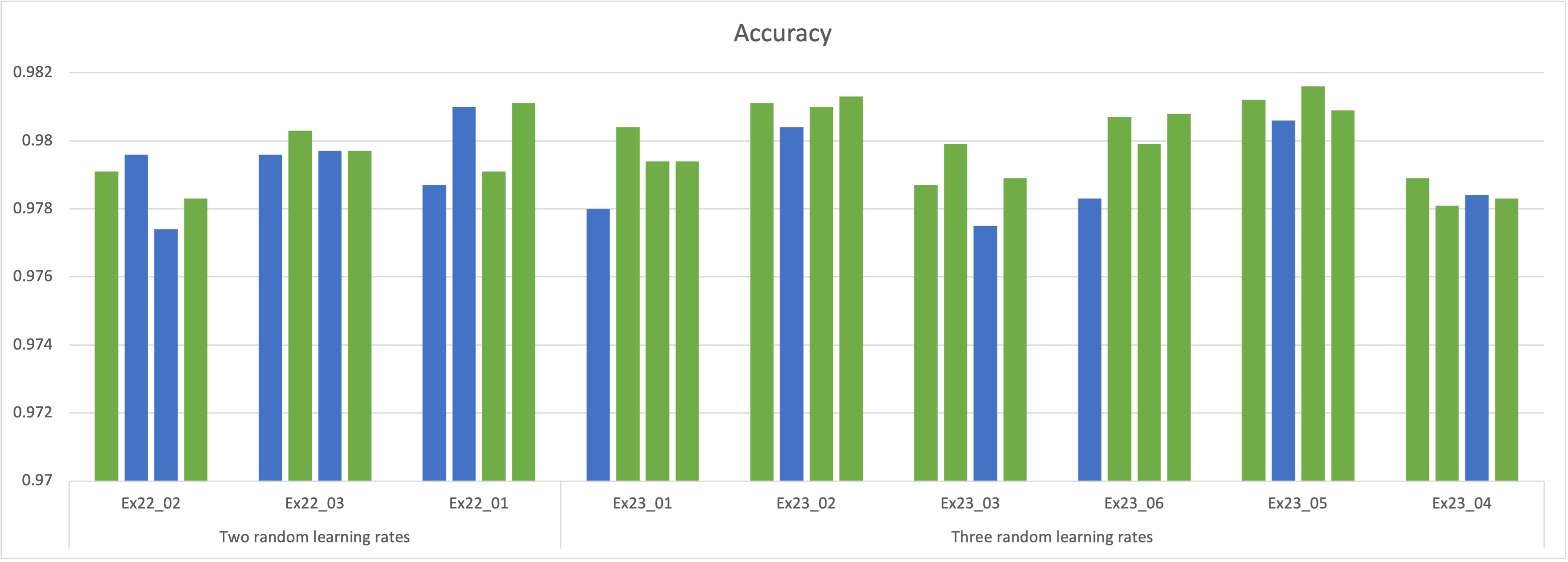}
\caption{Multi-random learning rates. The accuracy for experiments with two and three random particles. The latter is tested in two different settings. In each experiment, the columns from left to right hand side indicate PSO-1, PSO-2, PSO-3 and PSO-4 in order. When PSO-1, PSO-2 and PSO-3 are conservative the learning rates are set at $10^{-2}$, $10^{-3}$ and $10^{-4}$, respectively. The PSO-4’s learning rate is always in a random range. Green columns denote random PSOs whereas blue columns represent conservative ones.}
\label{fig:multirandom}
\end{center}
\end{figure*}

As we can see from Figure~\ref{fig:multirandom}, in the case of two wilder particles, a combination of two conservative PSO-1 and PSO-2 with larger learning rates ($10^{-2}$, $10^{-3}$) obtains a higher accuracy than the other options.

Regarding three wilder particles and one conservative particle, when PSO-2 is conservative (blue column), the group accomplishes a better performance than all the others in both settings. The rationale is that the conservative one provides an \emph{essential direction} for all the other three. In case the learning rate is small ($10^{-4}$), the particle slowly explores the solution space that affects the performance of the whole group. On the other hand, when the learning rate is higher ($10^{-2}$), the particle explores faster but may skip deeper minima, just discarding opportunities to improve accuracy. In other words, a conservative needs neither training slow nor fast to provide safer solutions at the times when other particles scan the landscape wildly. Therefore, conservative particle also plays an important role to the swarm even wilder particles are the key for improvement.

In general, experiments with three random learning rate outperform the experiments with two random learning rate ($0.9816$ versus $0.9811$).
%
%
\subsubsection{Cluster Warmup Learning Rate}
\begin{figure*}[htb!]
\begin{center}
\includegraphics[keepaspectratio,width=0.85\textwidth]{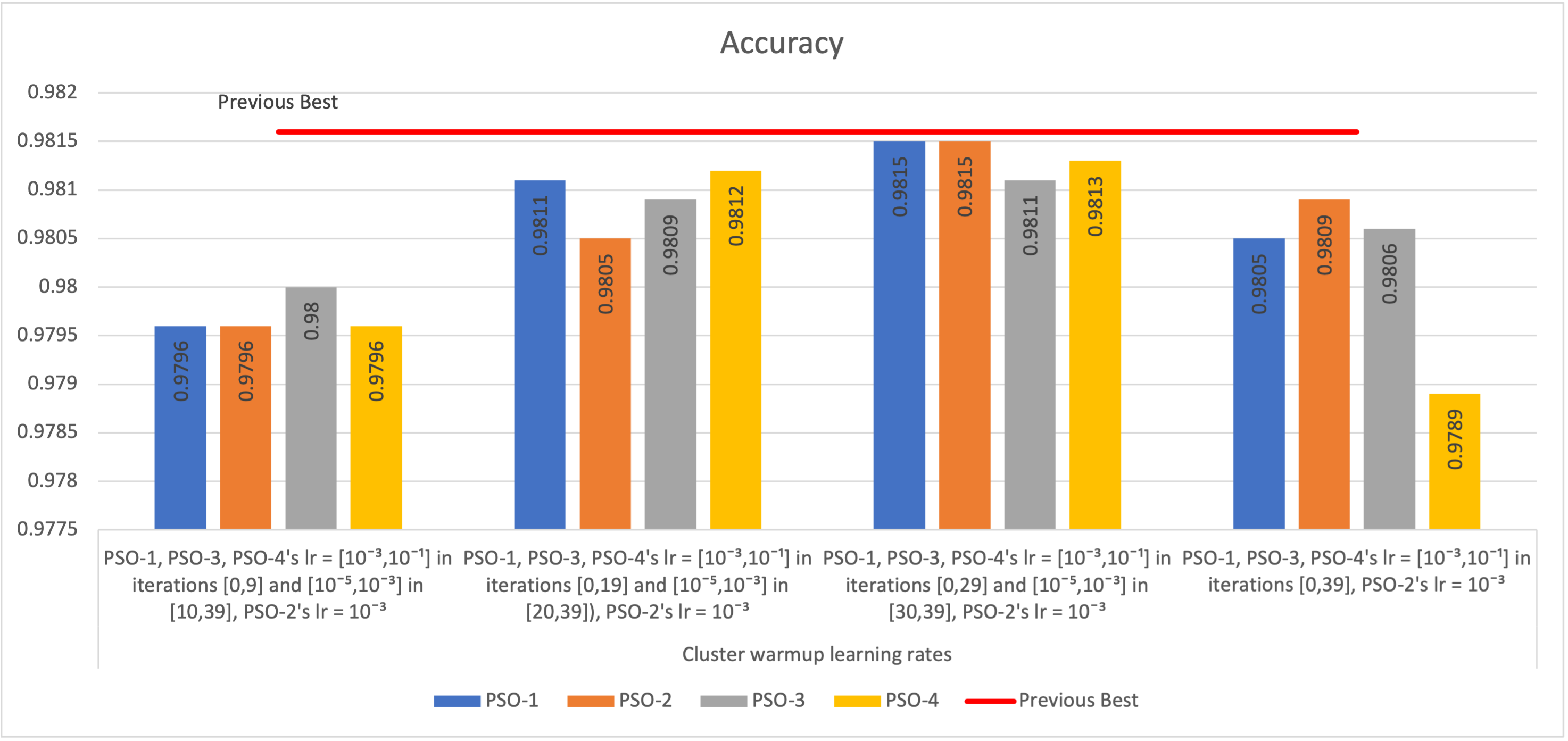}
\caption{Results for cluster warmup learning rates.}
\label{fig:cluster_wlr}
\end{center}
\end{figure*}

The results for these experiments are shown on Figure~\ref{fig:cluster_wlr}. Analogous to the typical training where the learning rate for a ConvNet is gradually reduced, we also setup a cluster of particles where learning rate range is reduced from the range [$10^{-3},10^{-1}$] to the range [$10^{-5},10^{-1}$] at distinct iterations, i.e., three PSOs have the random learning rate range changing and PSO-2 is fixed at $10^{-3}$ (the best setting obtained in the previous Section). Among choices, reducing learning rate after 30 iterations of a total of 40 yields a better accuracy ($0.9815$) despite slightly lower than the best result obtained before (see Section~\ref{sec:M}).
\subsubsection{Dynamics~$2$}
\label{sec:d2}
\begin{figure*}[htb!]
\begin{center}
\includegraphics[keepaspectratio,width=0.85\textwidth]{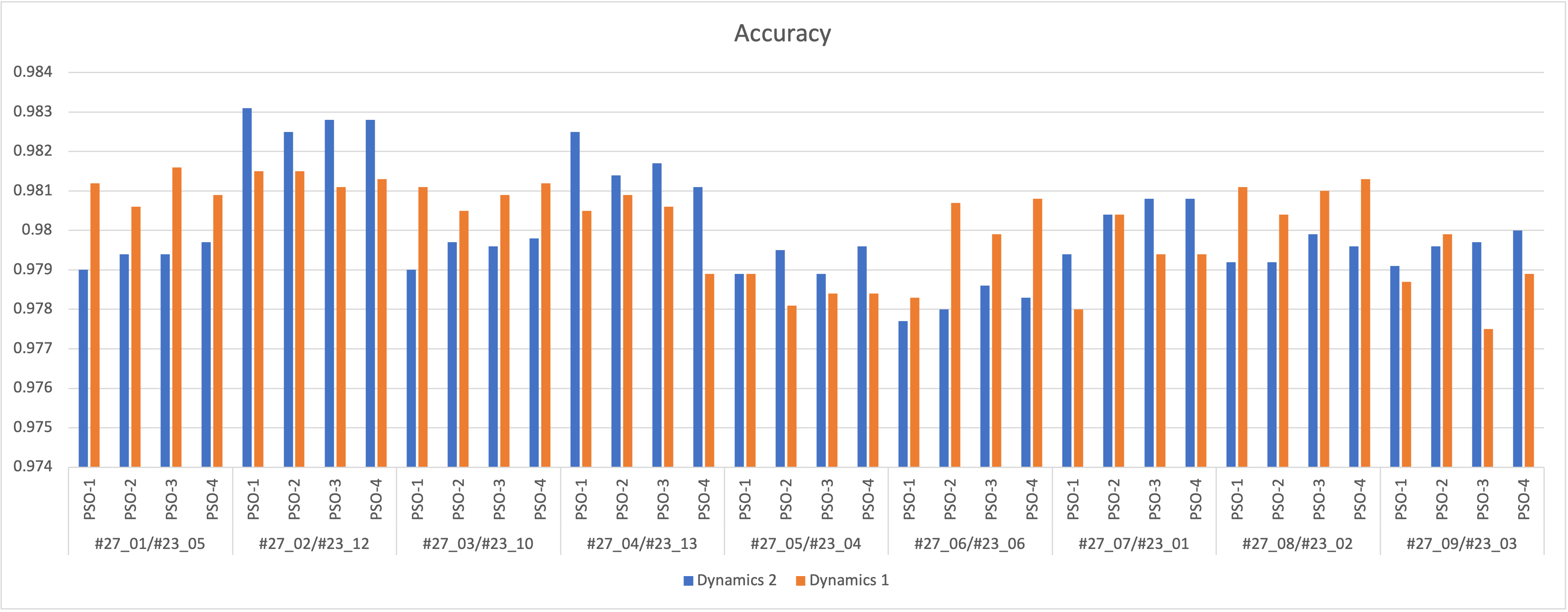}
\caption{Comparison of accuracy performance between Dynamics~$1$ and Dynamics~$2$. The latter outperforms the former and the best accuracy is further improved.}
\label{fig:f1f2}
\end{center}
\end{figure*}
In Section~\ref{sec:beta}, we have learnt that for some instances when a particle is coming closer to others, the particle tends to be pulled away. Theoretically, we would expect that, because of the effect of PSO algorithm, after a while particles will eventually stick together. Thus, these results seem to contradict our initial assumption (see equation~\eqref{eq:f1}, and therefore, we propose a modification to the algorithm so that the particle will be pulled back instead. As such the new formulation is stated in equation~\eqref{eq:f2} which we call Dynamics~$2$. 

We test the proposed algorithm in conjunction with strategies from previous sections and compare the performance of Dynamics~$2$ versus Dynamics~$1$. Figure~\ref{fig:f1f2} shows results of this comparison. We can see that, in most cases, Dynamics~$2$ not only outperforms the former, but, remarkably, achieved state-of-the-art result when the accuracy is improved to $0.9831$. This proves that the notion of pulling back particle is a critical mechanism for improvement.
\subsection{Performance on Cifar-100 benchmark dataset}
For a better evaluation of our proposed methods, apart from Cifar-10, we also performed experiments on the Cifar-100 benchmark dataset. In fact, as an example, we illustrated in Table~\ref{tab:table_performance_cifar100} the accuracy is increased by $2.78\%$ with Cifar-100 for Dynamics 2 with MobileNet confirming our paper claim.

\begin{figure}[htb!]
\begin{center}
\includegraphics[keepaspectratio,width=0.45\textwidth]{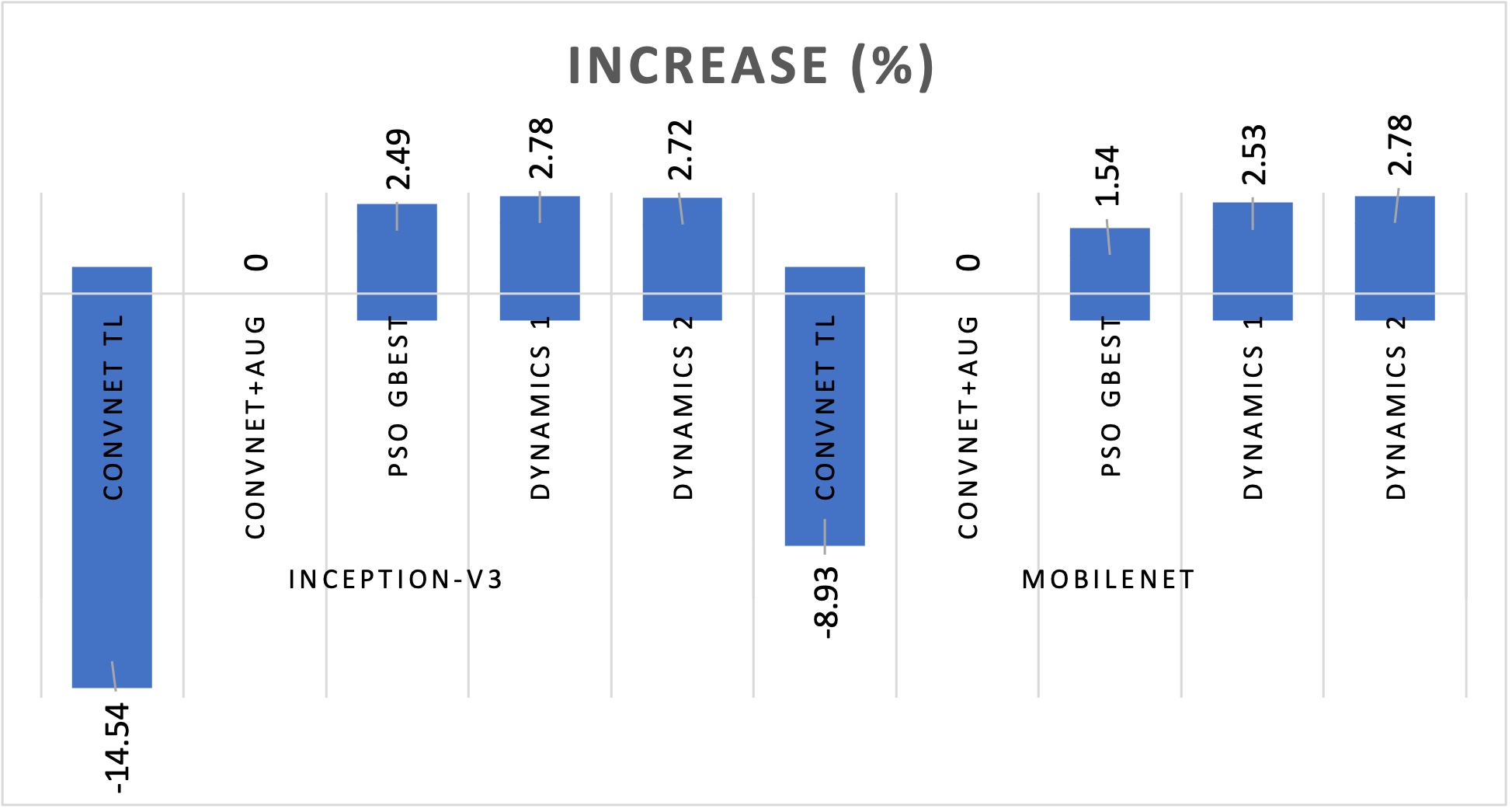}
\caption{Percentage increase of the proposed methods on Cifar-100.}
\label{fig:increase}
\end{center}
\end{figure}

These experiments were performed by running learning rate scans on Inception-v3 and MobileNet and the results indicate similar parameter ranges as in Cifar-10. For other settings, we reuse the best hyper-parameters which were found in the previous experiments. Even though, different datasets might require distinct settings but this strategy saves time using a rational guess. The results are illustrated in Figure~\ref{fig:increase} using the retrained architectures. The experiments confirm the effectiveness of our methods.

\input{table_performance_cifar100}
\section{Comparison with state-of-the-art methods}
\label{sec:stateoftheart}



The comparison results between the proposed method and other state-of-the-art counterpart algorithms are displayed in Table~\ref{tab:table_comparepsoconvnets}. A set of distinct methods were collected for a fair comparison. The results clearly show our algorithm outperforms other benchmark methods. We can see that ConvNets+Aug (fourth row counting from the bottom) performs $2.5\%$ and $0.75\%$ better than CLR and AmoebaNet in terms of accuracy on Cifar-10 dataset. In addition, this baseline method is much better than K-means and GA-CNN. Moreover, Tiled CNN method was achieved a lowest classification accuracy ($73.1\%$). Though, PyramidNet AMP outperforms our PSO gBest ($97.99\%$), the architecture is outweighed by Dynamics~$1$. Similarly, ResNet52-SAM performs better than Dynamics~$1$ but is surpassed by Dynamics~$2$ which yields $98.31\%$ accuracy. In the same manner, our proposed methods achieve the highest accuracy on Cifar-100 benchmark dataset, thus, confirming its superiority.

\input{table_comparepsoconvnets}
\section{Conclusion}
\label{sec:conclusion}
Image recognition has been a gold standard approach for many computer vision related tasks: extracting relevant information from telescope images in astronomy, navigation in robotics, cancer classification in medical image etc. However, training these large-scale neural networks for generalization is still a non-trivial task since the performance is sensitive to the architecture, the training set, the sample size among other attributes that renders the problem quite unstable to tuning. 

In this article, we have proposed a novel distributed collaborative PSO-ConvNet algorithm for the optimization performance which is capable of leading particles up to a better minimum. The key contributions of this article are: (1) novel formulations (Dynamics~$1$ and Dynamics~$2$) have been successfully created by incorporating distilled Cucker-Smale elements into the PSO algorithm using KNN and intertwining the training with SGD; (2) a new type of particle, i.e., wilder PSO with random learning rate is introduced which has capability of attracting conservative PSOs to stronger minima; (3) a distributed environment is developed for parallel collaboration that significantly accelerates the training; (4) the proposed algorithms are evaluated on Cifar-10 and Cifar-100 benchmark datasets and compared to state-of-the-art algorithms to verify the superior effectiveness. In the future, we will explore our algorithms on more datasets, e.g., ImageNet. In addition, we intend to incorporate our approach for action recognition.
\section*{Acknowledgments}
%
This research is sponsored by FEDER funds through the programs COMPETE -- ``Programa Operacional Factores de Competitividade'' and Centro2020 -- ``Centro Portugal Regional Operational Programme'', and by national funds through FCT -- ``Funda\c{c}\~{a}o para a Ci\^encia e a Tecnologia'', under the project UIDB/00326/2020. 
The support is gratefully acknowledged.
\bibliographystyle{unsrt}
\bibliography{hybridpso}
%
\clearpage
\begin{IEEEbiography}
[{\includegraphics[width=1in,height=1.2in,clip,keepaspectratio]{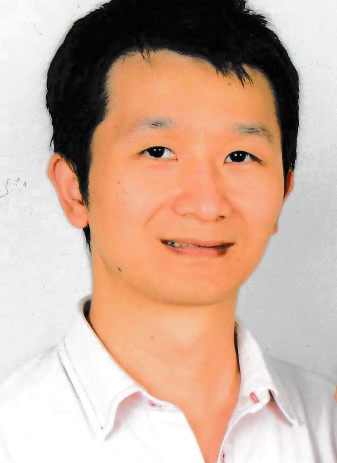}}]%
{Nguyen Huu Phong}
received the BSc in Physics from Vietnam National University, Hanoi and MSc degree in Information Technology from Shinawatra University, Thailand. He is currently a PhD student at the Department of Informatics Engineering of the University of Coimbra. He is member of the Adaptive Computation group at the Centre of Informatics and Systems at the University of Coimbra (CISUC). His research interests include image recognition, action recognition, pattern recognition, machine learning and deep learning.
\end{IEEEbiography}
\begin{IEEEbiography}
[{\includegraphics[width=1in,height=1.25in,clip,keepaspectratio]{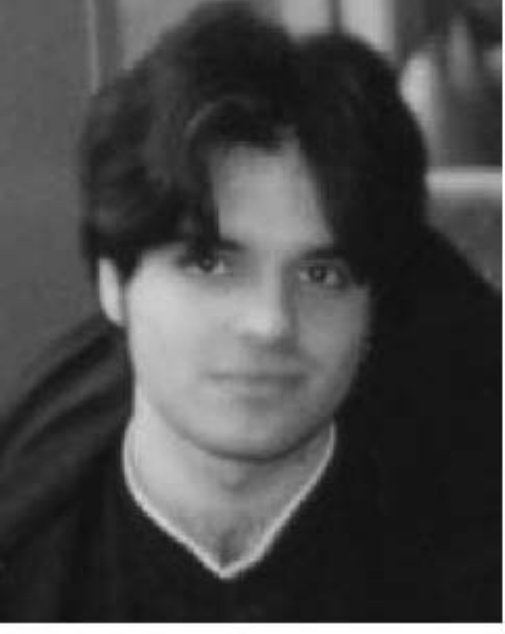}}]%
{Augusto Santos}
obtained his BSc. and MSc. from Instituto Superior Técnico (IST), Lisbon-Portugal, and Ph.D from Carnegie Mellon University, Pittsburgh-USA, and IST all in Electrical and Computer Engineering. He held a 2-year Postdoctoral scholar position at Carnegie Mellon University and another 2-year at the École Polytechnique Fédérale de Lausanne (EPFL), Lausanne-Switzerland. His current research concentrates on: i) high-dimensional statistical inference, especially on the problem of graph learning with latent variables; and ii) qualitative analysis of complex Networked Dynamical Systems. He is currently a researcher at the Centre for Informatics and Systems at the University of Coimbra (CISUC).
\end{IEEEbiography}
\begin{IEEEbiography}
[{\includegraphics[width=1in,height=1.2in,clip,keepaspectratio]{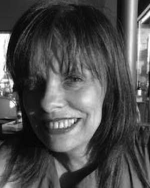}}]%
{Bernardete Ribeiro}
Bernardete Ribeiro (SM'15) received the Ph.D. in informatics engineering from University of Coimbra. She is a Full Professor with University of Coimbra and the former Director of the Center of Informatics and Systems of the University of Coimbra (CISUC). She is a senior member of Adaptive Computation group from CISUC and also the Founder and the Director of the Laboratory of  Neural Networks for over 25 years. Her research interests are in the areas of machine learning, and pattern recognition and their applications to a broad range of fields. She is the former President of the Portuguese Association of Pattern Recognition. She has been responsible/participated in several research projects both in international and national levels.  She is Associate Editor of IEEE Transactions on Cybernetics, IEEE SMC Senior Member and member of Association for Computing Machinery (ACM).
\end{IEEEbiography}
%
\EOD
\end{document}

%% file: table_initial_experiments.tex
\begin{table}
\centering
\caption{Accuracy on Cifar-10 for the Dynamics 1. Effects of KNN and Gradient are evaluated separately besides combination of the two. Results with pBest are also included. In all experiments, $c_2$ is kept at $0.5$}
\label{tab:table_basic_experiments}
\resizebox{\columnwidth}{!}{
\begin{tabular}{|l|l|l|l|l|c|} 
\hline
\multicolumn{1}{|c|}{Method}               & \multicolumn{4}{c|}{Accuracy}     & Index     \\ 
\cline{2-5}
                                           & PSO-1  & PSO-2  & PSO-3  & PSO-4  &           \\ 
\hline
KNN ($k=1$) & 0.9783 & 0.9717 & 0.9717 & \textbf{0.9785} & Ex17\_01  \\
KNN ($k=2$) & 0.9715 & 0.9759 & 0.9760 & 0.9769 & Ex17\_02  \\
KNN ($k=3$) & 0.9713 & 0.9773 & 0.9769 & 0.9765 & Ex17\_03  \\ 
\hline
Gradient ($M=10$)                            & 0.9750 & \textbf{0.9777} & 0.9708 & 0.9752 & Ex18\_01  \\
Gradient ($M=1$)                             & 0.9762 & 0.9730 & 0.9419 & 0.9699 & Ex18\_02  \\
Gradient ($M=0.1$)                           & 0.9744 & 0.9738 & 0.9343 & 0.9747 & Ex18\_03  \\ 
\hline
KNN ($k=1$), Gradient ($M=10$)                 & 0.9720 & 0.9755 & 0.9729 & 0.9754 & Ex19\_01  \\
KNN ($k=1$), Gradient ($M=1$)                  & 0.9796 & 0.9707 & 0.9712 & \textbf{0.9797} & Ex19\_02  \\
KNN ($k=3$), Gradient ($M=1$)                  & 0.9738 & 0.9790 & 0.9790 & 0.9787 & Ex19\_03  \\
\hline
KNN ($k=3$), Gradient ($M=2$), pBest ($c_1=0.5$)  & 0.9734 & 0.9783 & 0.9781 & 0.9774 & Ex13\_01  \\
KNN ($k=3$), Gradient ($M=2$), pBest ($c_1=0.2$)  & 0.9733 & \textbf{0.9785} & 0.9784 & 0.9777 & Ex13\_02  \\
\hline

\end{tabular}
}
\end{table}

%% file: table_adv_experiments.tex
\begin{table*}
\centering
\caption{Accuracy of PSOs on variety of $M$, $c_2$ and $warmup$ parameters}
\label{tab:gradients1}
\adjustbox{max width=\textwidth}{
\begin{tabular}{|c|c|c|c|c|c|c|c|c|c|c|c|c|c|c|c|c|c|c|} 
\hline
\multicolumn{15}{|c|}{\begin{tabular}[c]{@{}c@{}}\\Parameters\end{tabular}}                                                                                                                                                                                                                                                            & \multicolumn{4}{c|}{Accuracy}                                                                                                                                                      \\ 
\cline{1-15}
\multicolumn{2}{|c|}{KNN} & $warmup$ & \multicolumn{12}{c|}{Gradient}                                                                                                                                                                                                                                                                    & \multicolumn{4}{c|}{}                                                                                                                                                              \\ 
\cline{1-2}\cline{4-19}
$k$ & $c_2$                     &        & \multicolumn{3}{c|}{PSO-1}                                                  & \multicolumn{3}{c|}{PSO-2}                                                  & \multicolumn{3}{c|}{PSO-3}                                                  & \multicolumn{3}{c|}{PSO-4}                              & \multicolumn{1}{c}{PSO-1}                  & \multicolumn{1}{c}{PSO-2}                  & \multicolumn{1}{c}{PSO-3}                  & PSO-4                                       \\ 
\cline{4-15}
  &                       &        & \multicolumn{1}{c}{M12} & \multicolumn{1}{c}{M13} & \multicolumn{1}{c}{M14} & \multicolumn{1}{c}{M21} & \multicolumn{1}{c}{M23} & \multicolumn{1}{c}{M24} & \multicolumn{1}{c}{M31} & \multicolumn{1}{c}{M32} & \multicolumn{1}{c}{M34} & \multicolumn{1}{c}{M41} & \multicolumn{1}{c}{M42} & M43 & \multicolumn{1}{c}{}                       & \multicolumn{1}{c}{}                       & \multicolumn{1}{c}{}                       &                                             \\ 
\hline
3 & 1.7                   & 0      & 0.2                     & 0.2                     & 1.7                     & 1.7                     & 0.2                     & 1.7                     & 1.7                     & 1.7                     & 1.7                     & 1.7                     & 0.2                     & 0.2 & 0.9740                                     & 0.9797                                     & 0.9798                                     & 0.9801                                      \\ 
\hline
3 & 1.7                   & 10     & 0.2                     & 0.2                     & 1.7                     & 1.7                     & 0.2                     & 1.7                     & 1.7                     & 1.7                     & 1.7                     & 1.7                     & 0.2                     & 0.2 & 0.9756                                     & 0.9803                                     & 0.9798                                     & 0.9802                                      \\ 
\hline
3 & 1.7                   & 20     & 0.2                     & 0.2                     & 1.7                     & 1.7                     & 0.2                     & 1.7                     & 1.7                     & 1.7                     & 1.7                     & 1.7                     & 0.2                     & 0.2 & 0.9743                                     & 0.9798                                     & 0.9801                                     & 0.9796                                      \\ 
\hline
3 & 1.7                   & 30     & 0.2                     & 0.2                     & 1.7                     & 1.7                     & 0.2                     & 1.7                     & 1.7                     & 1.7                     & 1.7                     & 1.7                     & 0.2                     & 0.2 & 0.9741                                     & 0.9778                                     & 0.9772                                     & 0.9773                                      \\ 
\hline
3 & 1.2                   & 0      & 0.2                     & 0.2                     & 1.7                     & 1.7                     & 0.2                     & 1.7                     & 1.7                     & 1.7                     & 1.7                     & 1.7                     & 0.2                     & 0.2 & \textcolor[rgb]{0.122,0.122,0.129}{0.9779} & \textcolor[rgb]{0.122,0.122,0.129}{0.9713} & \textcolor[rgb]{0.122,0.122,0.129}{0.9783} & \textcolor[rgb]{0.122,0.122,0.129}{0.9782}  \\ 
\hline
3 & 1.2                   & 10     & 0.2                     & 0.2                     & 1.7                     & 1.7                     & 0.2                     & 1.7                     & 1.7                     & 1.7                     & 1.7                     & 1.7                     & 0.2                     & 0.2 & 0.9758                                     & 0.9789                                     & 0.9789                                     & 0.9792                                      \\ 
\hline
3 & 1.2                   & 20     & 0.2                     & 0.2                     & 1.7                     & 1.7                     & 0.2                     & 1.7                     & 1.7                     & 1.7                     & 1.7                     & 1.7                     & 0.2                     & 0.2 & 0.9719                                     & 0.9781                                     & 0.9778                                     & 0.9780                                      \\ 
\hline
3 & 1.2                   & 30     & 0.2                     & 0.2                     & 1.7                     & 1.7                     & 0.2                     & 1.7                     & 1.7                     & 1.7                     & 1.7                     & 1.7                     & 0.2                     & 0.2 & 0.9759                                     & 0.9791                                     & 0.9793                                     & 0.9781                                      \\ 
\hline
3 & 0.2                   & 0      & 0.2                     & 0.2                     & 1.7                     & 1.7                     & 0.2                     & 1.7                     & 1.7                     & 1.7                     & 1.7                     & 1.7                     & 0.2                     & 0.2 & \textcolor[rgb]{0.122,0.122,0.129}{0.9764} & \textcolor[rgb]{0.122,0.122,0.129}{0.9798} & \textcolor[rgb]{0.122,0.122,0.129}{0.9798} & \textcolor[rgb]{0.122,0.122,0.129}{0.9788}  \\ 
\hline
3 & 0.2                   & 10     & 0.2                     & 0.2                     & 1.7                     & 1.7                     & 0.2                     & 1.7                     & 1.7                     & 1.7                     & 1.7                     & 1.7                     & 0.2                     & 0.2 & \textcolor[rgb]{0.122,0.122,0.129}{0.9746} & \textcolor[rgb]{0.122,0.122,0.129}{0.9777} & \textcolor[rgb]{0.122,0.122,0.129}{0.9781} & \textcolor[rgb]{0.122,0.122,0.129}{0.9788}  \\ 
\hline
3 & 0.2                   & 20     & 0.2                     & 0.2                     & 1.7                     & 1.7                     & 0.2                     & 1.7                     & 1.7                     & 1.7                     & 1.7                     & 1.7                     & 0.2                     & 0.2 & \textcolor[rgb]{0.122,0.122,0.129}{0.9761} & \textcolor[rgb]{0.122,0.122,0.129}{0.9786} & \textcolor[rgb]{0.122,0.122,0.129}{0.9795} & \textcolor[rgb]{0.122,0.122,0.129}{0.9774}  \\ 
\hline
3 & 0.2                   & 30     & 0.2                     & 0.2                     & 1.7                     & 1.7                     & 0.2                     & 1.7                     & 1.7                     & 1.7                     & 1.7                     & 1.7                     & 0.2                     & 0.2 & \textcolor[rgb]{0.122,0.122,0.129}{0.9768} & \textcolor[rgb]{0.122,0.122,0.129}{0.9770} & \textcolor[rgb]{0.122,0.122,0.129}{0.9771} & \textcolor[rgb]{0.122,0.122,0.129}{0.9760}  \\ 
\hline
3 & 0.5                   & 0      & 0.2                     & 0.2                     & 1.7                     & 1.7                     & 0.2                     & 1.7                     & 1.7                     & 1.7                     & 1.7                     & 1.7                     & 0.2                     & 0.2 & \textcolor[rgb]{0.122,0.122,0.129}{0.9773} & \textcolor[rgb]{0.122,0.122,0.129}{0.9802} & \textcolor[rgb]{0.122,0.122,0.129}{0.9800} & \textcolor[rgb]{0.122,0.122,0.129}{0.9797}  \\ 
\hline
3 & 0.5                   & 10     & 0.2                     & 0.2                     & 1.7                     & 1.7                     & 0.2                     & 1.7                     & 1.7                     & 1.7                     & 1.7                     & 1.7                     & 0.2                     & 0.2 & \textcolor[rgb]{0.122,0.122,0.129}{0.9744} & \textcolor[rgb]{0.122,0.122,0.129}{0.9789} & \textcolor[rgb]{0.122,0.122,0.129}{0.9792} & \textcolor[rgb]{0.122,0.122,0.129}{0.9788}  \\ 
\hline
3 & 0.5                   & 20     & 0.2                     & 0.2                     & 1.7                     & 1.7                     & 0.2                     & 1.7                     & 1.7                     & 1.7                     & 1.7                     & 1.7                     & 0.2                     & 0.2 & \textcolor[rgb]{0.122,0.122,0.129}{0.9716} & \textcolor[rgb]{0.122,0.122,0.129}{0.9786} & \textcolor[rgb]{0.122,0.122,0.129}{0.9787} & \textcolor[rgb]{0.122,0.122,0.129}{0.9782}  \\ 
\hline
3 & 0.5                   & 30     & 0.2                     & 0.2                     & 1.7                     & 1.7                     & 0.2                     & 1.7                     & 1.7                     & 1.7                     & 1.7                     & 1.7                     & 0.2                     & 0.2 & \textcolor[rgb]{0.122,0.122,0.129}{0.9758} & \textcolor[rgb]{0.122,0.122,0.129}{0.9783} & \textcolor[rgb]{0.122,0.122,0.129}{0.9788} & \textcolor[rgb]{0.122,0.122,0.129}{0.9777}  \\ 
\hline
3 & 0.5                   & 0      & 0.2                     & 0.2                     & 10                      & 1.7                     & 0.2                     & 10                      & 1.7                     & 1.7                     & 10                      & 1.7                     & 0.2                     & 0.2 & \textcolor[rgb]{0.122,0.122,0.129}{0.9774} & \textcolor[rgb]{0.122,0.122,0.129}{0.9814} & \textcolor[rgb]{0.122,0.122,0.129}{0.9814} & \textcolor[rgb]{0.122,0.122,0.129}{0.9816}  \\ 
\hline
3 & 0.5                   & 10     & 0.2                     & 0.2                     & 10                      & 1.7                     & 0.2                     & 10                      & 1.7                     & 1.7                     & 10                      & 1.7                     & 0.2                     & 0.2 & \textcolor[rgb]{0.122,0.122,0.129}{0.9764} & \textcolor[rgb]{0.122,0.122,0.129}{0.9806} & \textcolor[rgb]{0.122,0.122,0.129}{0.9811} & \textcolor[rgb]{0.122,0.122,0.129}{0.9805}  \\ 
\hline
3 & 0.5                   & 20     & 0.2                     & 0.2                     & 10                      & 1.7                     & 0.2                     & 10                      & 1.7                     & 1.7                     & 10                      & 1.7                     & 0.2                     & 0.2 & \textcolor[rgb]{0.122,0.122,0.129}{0.9770} & \textcolor[rgb]{0.122,0.122,0.129}{0.9782} & \textcolor[rgb]{0.122,0.122,0.129}{0.9782} & \textcolor[rgb]{0.122,0.122,0.129}{0.9775}  \\
\hline
\end{tabular}
}
\end{table*}

%% file: table_settings.tex
\begin{table}
\centering
\caption{Settings weight $M$}
\label{tab:gradients}
\resizebox{\columnwidth}{!}{
\begin{tabular}{|c|l|l|l|l|l|l|l|l|l|l|l|} 
\cline{2-12}
\multicolumn{1}{l|}{} & \multicolumn{11}{c|}{Gradient Weights}                                                                                                                                                                                                                                                                  \\ 
\hline
Settings              & \multicolumn{1}{c|}{M12} & \multicolumn{1}{c|}{M13} & \multicolumn{1}{c|}{M14} & \multicolumn{1}{c|}{M21} & \multicolumn{1}{c|}{M23} & \multicolumn{1}{c|}{M24} & \multicolumn{1}{c|}{M31} & \multicolumn{1}{c|}{M32} & \multicolumn{1}{c|}{M34} & \multicolumn{1}{c|}{M42} & \multicolumn{1}{c|}{M43}  \\ 
\hline
S1                    & 0.2                      & 0.2                      & 10                       & 1.7                      & 0.2                      & 10                       & 1.7                      & 1.7                      & 10                       & 1.7                      & 0.2                       \\ 
\hline
S2                    & 0.2                      & 0.2                      & 1.7                      & 1.7                      & 0.2                      & 1.7                      & 1.7                      & 1.7                      & 1.7                      & 1.7                      & 0.2                       \\
\hline
\end{tabular}
}
\end{table}

%% file: table_parameters.tex
\begin{table}[htb!]
\centering
\caption{\label{tab:table_parameters}ConvNets Hyper-parameters}
\begin{tabular}{lc} 
\hline
                                  & \multicolumn{1}{l}{}         \\
ConvNets                 &                              \\
\# of iterations full training    & 40                           \\
Gaussian noise standard deviation & 0.1                          \\
Fully connected number of neurons & 1024                         \\
Batch size                        & 32                           \\
                                  & \multicolumn{1}{l}{}         \\
Augmentation                      & \multicolumn{1}{l}{}         \\
Rotation range                    & 30                           \\
Width shift range                 & 0.3                          \\
Height shift range                & 0.3                          \\
Shear range                       & 0.3                          \\
Zoom range                        & 0.3                          \\
Channel shift range               & 10                           \\
Fill mode                         & \multicolumn{1}{l}{nearest}  \\
Preprocessing                     & {[}-1 1]                     \\
\hline
\end{tabular}
\end{table}

%% file: table_performance_cifar100.tex
\begin{table}[htb!]
\centering
\caption{Performance on Cifar-100. The methods ConvNet TL (Transfer Learning), ConvNet+Aug (Augmentation), PSO gBest, Dynamics 1 and Dynamics 2 are evaluated on retrained architectures Inception-v3 and MobileNet. The ConvNet+Aug method is utilized as a baseline. The most accuracy among PSOs is selected to compute the percentage increase.}
\label{tab:table_performance_cifar100}
\resizebox{\columnwidth}{!}{
\begin{tabular}{|l|l|l|l|l|l|r|} 
\hline
\multicolumn{1}{|c|}{\multirow{2}{*}{Architecture}} & \multicolumn{1}{c|}{\multirow{2}{*}{Method}} & \multicolumn{4}{c|}{Accuracy}                                                                                     & \multirow{2}{*}{Increase (\%)}  \\ 
\cline{3-6}
\multicolumn{1}{|c|}{}                              & \multicolumn{1}{c|}{}                        & \multicolumn{1}{c|}{PSO-1} & \multicolumn{1}{c|}{PSO-2} & \multicolumn{1}{c|}{PSO-3} & \multicolumn{1}{c|}{PSO-4} &                                 \\ 
\hline
\multirow{5}{*}{Inception-v3}                       & ConvNet TL                                     & 0.7014                     & 0.6714                     & 0.5309                     & 0.7016                     & -14.54                          \\ 
\cline{2-7}
                                                    & ConvNet+Aug                                      & 0.8454                     & 0.847                      & 0.6725                     & 0.8428                     & 0                             \\ 
\cline{2-7}
                                                    & PSO gBest                                    & 0.8471                     & 0.8719                     & 0.865                      & 0.8676                     & +2.49                            \\ 
\cline{2-7}
                                                    & Dynamics 1                                   & 0.8729                     & 0.8736                     & 0.8744                     & 0.8748                     & +2.78                            \\ 
\cline{2-7}
                                                    & Dynamics 2                                   & 0.8742                     & 0.8709                     & 0.8724                     & 0.8734                     & +2.72                            \\ 
\hline
\multirow{5}{*}{MobileNet}                          & ConvNet TL                                     & 0.7076                     & 0.6951                     & 0.6786                     & 0.6988                     & -8.93                           \\ 
\cline{2-7}
                                                    & ConvNet+Aug                                      & 0.7791                     & 0.7969                     & 0.7693                     & 0.7951                     & 0                             \\ 
\cline{2-7}
                                                    & PSO gBest                                    & 0.7699                     & 0.8119                     & 0.8106                     & 0.8123                     & +1.54                           \\ 
\cline{2-7}
                                                    & Dynamics 1                                   & 0.8216                     & 0.813                      & 0.8222                     & 0.8215                     & +2.53                           \\ 
\cline{2-7}
                                                    & Dynamics 2                                   & 0.8243                     & 0.8158                     & 0.8246                     & 0.8247                     & \textbf{+2.78}                           \\
\hline
\end{tabular}
}
\end{table}

%% file: table_comparepsoconvnets.tex
\begin{table}[hbt!]
\centering
\caption{Classification accuracy (\%) on the Cifar-10 and Cifar-100 benchmark datasets.}
\label{tab:table_comparepsoconvnets}
\makebox[\linewidth]{
\begin{tabular}{|l|c|c|} 
\hline
\multirow{2}{*}{Method}  & \multicolumn{2}{c|}{Accuracy (\%)}  \\ 
\cline{2-3}
                         & Cifar-10       & Cifar-100          \\ 
\hline
Tiled CNN~ \cite{ngiam2010tiled}              & 73.10          & -                  \\ 
\hline
K-means~\cite{coates2011analysis}                 & 79.64          & -                  \\ 
\hline
GA-CNN~\cite{young2015optimizing}                  & 74.59          & -                  \\ 
\hline
cPSO-CNN~\cite{wang2019cpso}                     & 91.35          & -              \\ 
\hline
CLR~\cite{smith2017cyclical}                     & 94.90          & 75.90              \\ 
\hline
AmoebaNet-A (N=6, F=36)~\cite{real2019regularized} & 96.66          & -                  \\ 
\hline
ENAS~\cite{pham2018efficient}                    & 97.11          & -                  \\ 
\hline
SENet~\cite{hu2018squeeze}                   & 97.88          & 84.59              \\ 
\hline
PyramidNet AMP~\cite{zheng2021regularizing}          & 98.02          & 86.64              \\ 
\hline
ViT-B/16~\cite{dosovitskiy2020image}                & 98.13          & 87.13              \\ 
\hline
ResNet152-SAM~\cite{chen2021vision}           & 98.20          & 87.08              \\ 
\hhline{|===|}
ConvNet+Aug         & 97.41          & 84.7               \\ 
\hline
PSO gBest             & 97.99          & 87.19              \\ 
\hline
Dynamics 1               & 98.16          & \textbf{87.48}     \\ 
\hline
Dynamics 2               & \textbf{98.31} & 87.42              \\
\hline
\end{tabular}
}
\begin{tablenotes}
    \small
    \item The bold letter indicates that the proposed algorithm achieved best efficiency amongst the compared algorithms
\end{tablenotes}
\end{table}